\documentclass[11pt, a4paper]{lumia}

\usepackage[sort&compress]{natbib}
\usepackage{xltabular}
\usepackage{arydshln}
\usepackage{longtable}

\usepackage{algorithm}
\newcommand{\KwIn}[1]{\item[\textbf{Input:}] #1}
\newcommand{\KwOut}[1]{\item[\textbf{Output:}] #1}
\usepackage{algpseudocode} 
\usepackage{amssymb}
\usepackage{amsmath}
\usepackage{wrapfig}
\definecolor{commentcolor}{HTML}{067f7f}
\newcommand{\imp}[1]{{\small\color{red}(+#1)}}

\usepackage[most,skins,theorems]{tcolorbox} 
\usepackage[commandnameprefix=always]{changes}

\newtcolorbox{promptbox}[2][]{
  colback=gray!5,       
  colframe=black!70,    
  title=\textbf{#2},     
  fonttitle=\bfseries\small,
  fontupper=\footnotesize\ttfamily, 
  left=3pt, right=3pt, top=3pt, bottom=3pt, 
  boxrule=0.8pt,
  sharp corners,         
  #1
}

\setheadertext{\texttt{FlowPIE}: Test-Time Scientific Idea Evolution with Flow-Guided Literature Exploration}

\renewcommand{\today}{2026-04}

\usepackage{xspace}

\theoremstyle{plain}

\newtheorem*{proposition*}{Proposition}

\theoremstyle{definition}

\theoremstyle{definition}

\def\eqref#1{equation~\ref{#1}}

\usepackage[utf8]{inputenc}     
\usepackage[T1]{fontenc}        
\usepackage{microtype}          

\usepackage{xcolor}             
\usepackage[dvipsnames]{xcolor} 
\usepackage{graphicx}           
\usepackage{tikz}               
\usepackage[edges]{forest}      

\usepackage{hyperref}           
\usepackage{url}                
\usepackage{xurl}               

\usepackage{tabularx}           
\usepackage{array}              
\usepackage{booktabs}           
\usepackage{longtable}          
\usepackage{multirow}           
\usepackage{makecell}           
\usepackage[table]{xcolor}      
\usepackage{colortbl}           
\usepackage{ragged2e}           

\usepackage{amsmath}            
\usepackage{mathtools}          
\usepackage{amsfonts}           
\usepackage{amssymb}            
\usepackage{amsthm}             
\usepackage{nicefrac}           

\usepackage{algorithm}          
\usepackage{algorithmicx}       
\usepackage{algpseudocode}      
\usepackage{listings}           

\usepackage{subcaption}         
\usepackage{caption}            
\usepackage{wrapfig}            
\usepackage[export]{adjustbox}  

\usepackage{soul}               
\usepackage{changes}            
\usepackage{xspace}             
\usepackage[normalem]{ulem}     
\usepackage{CJKutf8}            

\usepackage{enumitem}           
\usepackage{pifont}             

\usepackage[tikz]{bclogo}       
\usepackage[framemethod=tikz]{mdframed} 

\usepackage{lipsum}             
\usepackage{tocloft}            
\usepackage{fontawesome5}       
\usepackage{afterpage}          
\usepackage{bbding}             
\usepackage{epigraph}           
\usepackage{minitoc}            
\usepackage{multicol}           
\usepackage{textgreek}          

\newcolumntype{P}[1]{>{\RaggedRight\arraybackslash}p{#1}}

\definecolor{darkblue}{rgb}{0, 0, 0.5}
\definecolor{uclablue}{RGB}{39, 116, 174}
\definecolor{bigaired}{RGB}{156, 0, 0}
\definecolor{myblue}{HTML}{598BE7}
\definecolor{mildblue}{RGB}{31,119,180}
\definecolor{sectionblue}{RGB}{70, 130, 180}
\definecolor{methodblue}{RGB}{0, 150, 136}
\definecolor{bgblue}{RGB}{245,243,253}
\definecolor{ttblue}{RGB}{91,194,224}
\definecolor{mygreen}{rgb}{0.64, 0.56, 0.88}
\definecolor{myyellow}{rgb}{0.68, 0.6, 0.1}
\definecolor{fancygreen}{rgb}{0.33, 0.68, 0.20}
\definecolor{salmon}{rgb}{0.94, 0.52, 0.49}
\definecolor{tablegreen}{rgb}{0.82, 0.94, 0.75}
\definecolor{tableblue}{rgb}{0.81, 0.90, 0.94}
\definecolor{tablered}{rgb}{0.97, 0.85, 0.85}
\definecolor{tableorange}{rgb}{0.96, 0.85, 0.81}
\definecolor{myorange}{rgb}{1.0, 0.49, 0.0}
\definecolor{tlgreen}{rgb}{0.33, 0.68, 0.20}
\definecolor{darkgreen}{RGB}{0,100,0}
\definecolor{darkred}{RGB}{200, 0, 0}
\definecolor{lightblue}{RGB}{220,235,250}
\definecolor{customyellow}{HTML}{FFFACD}
\definecolor{refinegreen}{RGB}{0, 128, 75}
\definecolor{scoregreen}{RGB}{34, 139, 34}
\definecolor{hidden-blue}{RGB}{194,232,247}
\definecolor{hidden-black}{RGB}{20,68,106}
\definecolor{yes}{HTML}{C6EFCE}
\definecolor{no}{HTML}{FFC7CE}
\definecolor{partial}{HTML}{FFEB9C}
\definecolor{external}{HTML}{D9E1F2}
\definecolor{hdr}{HTML}{F2F2F2}
\definecolor{GRPOrow}{gray}{0.96}
\definecolor{FlowRLrow}{RGB}{225,236,255}
\definecolor{FlowBlue}{RGB}{80,120,210}
\definecolor{GRPOGray}{gray}{0.35}

\hypersetup{
    colorlinks=true, 
    citecolor=uclablue, 
    linkcolor=bigaired,
    urlcolor=darkblue
}

\setlist[itemize]{leftmargin=20pt, noitemsep, topsep=0pt}

\NewDocumentCommand{\kaiyan}{mO{}}{\textcolor{purple}{\textsuperscript{\textit{kaiyan}}\textsf{\textbf{\small[#1]}}}}
\NewDocumentCommand{\yuxin}{mO{}}{\textcolor{cyan}{\textsuperscript{\textit{yuxin}}\textsf{\textbf{\small[#1]}}}}
\NewDocumentCommand{\bx}{mO{}}{\textcolor{green}{\textsuperscript{\textit{bx}}\textsf{\textbf{\small[#1]}}}}
\NewDocumentCommand{\at}{mO{}}{\textcolor{red}{\textsuperscript{\textit{AT}}\textsf{\textbf{\small[#1]}}}}
\NewDocumentCommand{\re}{mO{}}{\textcolor{blue}{\textsuperscript{\textit{RE}}\textsf{\textbf{\small[#1]}}}}
\NewDocumentCommand{\ybsun}{mO{}}{\textcolor{magenta}{\textsuperscript{\textit{youbang}}\textsf{\textbf{\small[#1]}}}}
\NewDocumentCommand{\runze}{mO{}}{\textcolor{orange}{\textsuperscript{\textit{runze}}\textsf{\textbf{\small[#1]}}}}
\NewDocumentCommand{\add}{mO{}}{\textcolor{darkgreen}{\textsuperscript{\textit{Maybe Consider Discuss}}\textsf{\textbf{[#1]}}}}

\newcommand{\cmark}{\textcolor{darkgreen}{\boldmath$\checkmark$}}
\newcommand{\xmark}{\textcolor{darkred}{\boldmath$\times$}}

\newenvironment{itemize*}%
 {\leftmargini=10pt\begin{itemize}%
  \setlength{\itemsep}{0pt}%
  \setlength{\parskip}{0pt}%
  }%
 {\end{itemize}}

\newenvironment{enumerate*}%
 {\begin{enumerate}%
  \setlength{\itemsep}{0pt}%
  \setlength{\parskip}{0pt}}%
 {\end{enumerate}}

\newcommand{\cellstatus}[1]{%
  \begingroup
  \StrTrim{#1}[\statusval]%
  \IfStrEq{\statusval}{Yes}{\cellcolor{yes}\cmark}{}%
  \IfStrEq{\statusval}{No}{\cellcolor{no}\xmark}{}%
  \IfBeginWith{\statusval}{Yes (}{\cellcolor{yes}\cmark~\textit{\statusval\unskip}}{}%
  \IfStrEq{\statusval}{Partial}{\cellcolor{partial}\textbf{Partial}}{}%
  \IfStrEq{\statusval}{External}{\cellcolor{external}\textbf{External}}{}%
  \endgroup
}

\newtcolorbox{myboxi}[1][]{
  breakable,
  title=#1,
  colback=red!5,
  colbacktitle=red!5,
  coltitle=black,
  fonttitle=\bfseries,
  bottomrule=0pt,
  toprule=0pt,
  leftrule=2pt,
  rightrule=2pt,
  titlerule=0pt,
  arc=0pt,
  outer arc=0pt,
  colframe=red,
}

\newtcolorbox{myboxnote}[1][]{
  breakable,
  title=#1,
  colback=orange!0,
  colbacktitle=orange!0,
  coltitle=black,
  fonttitle=\bfseries,
  bottomrule=0pt,
  toprule=0pt,
  leftrule=2pt,
  rightrule=2pt,
  titlerule=0pt,
  arc=0pt,
  outer arc=0pt,
  colframe=orange,
}

\newtcolorbox{myboxii}[1][]{
  breakable,
  freelance,
  title=#1,
  colback=white,
  colbacktitle=white,
  coltitle=black,
  fonttitle=\bfseries,
  bottomrule=0pt,
  boxrule=0pt,
  colframe=white,
  overlay unbroken and first={
  \draw[red!75!black,line width=3pt]
    ([xshift=5pt]frame.north west) -- 
    (frame.north west) -- 
    (frame.south west);
  \draw[red!75!black,line width=3pt]
    ([xshift=-5pt]frame.north east) -- 
    (frame.north east) -- 
    (frame.south east);
  },
  overlay unbroken app={
  \draw[red!75!black,line width=3pt,line cap=rect]
    (frame.south west) -- 
    ([xshift=5pt]frame.south west);
  \draw[red!75!black,line width=3pt,line cap=rect]
    (frame.south east) -- 
    ([xshift=-5pt]frame.south east);
  },
  overlay middle and last={
  \draw[red!75!black,line width=3pt]
    (frame.north west) -- 
    (frame.south west);
  \draw[red!75!black,line width=3pt]
    (frame.north east) -- 
    (frame.south east);
  },
  overlay last app={
  \draw[red!75!black,line width=3pt,line cap=rect]
    (frame.south west) --
    ([xshift=5pt]frame.south west);
  \draw[red!75!black,line width=3pt,line cap=rect]
    (frame.south east) --
    ([xshift=-5pt]frame.south east);
  },
}

\tcbset{
  takeawaysbox/.style={
    title=Takeaways,
    colback=lightblue!80,
    colframe=black,
    fonttitle=\bfseries\small,
    coltitle=white,
    colbacktitle=black,
    enhanced,
    attach boxed title to top left={xshift=2.5mm,yshift=-2.5mm},
    boxed title style={rounded corners, size=small, colframe=black, colback=black},
    width=\linewidth,
    arc=3.5mm
  }
}

\mdfdefinestyle{mystyle}{%
  rightline=true,
  innerleftmargin=10,
  innerrightmargin=10,
  outerlinewidth=3pt,
  topline=false,
  rightline=true,
  bottomline=false,
  skipabove=\topsep,
  skipbelow=\topsep
}

\tikzset{%
    every node/.style={font=\tiny},
    parent/.style =          {align=center,text width=2cm,rounded corners=3pt, line width=0.3mm, fill=gray!10,draw=gray!80},
    child/.style =           {align=center,text width=2.0cm,rounded corners=3pt, fill=blue!10,draw=blue!80,line width=0.3mm},
    grandchild/.style =      {align=center,text width=2cm,rounded corners=3pt},
    greatgrandchild/.style = {align=center,text width=1.5cm,rounded corners=3pt},
    greatgrandchild2/.style = {align=center,text width=1.5cm,rounded corners=3pt},    
    referenceblock/.style =  {align=center,text width=1.5cm,rounded corners=2pt},
    pretrain/.style =           {align=center,text width=2.0cm,rounded corners=3pt, fill=blue!10,draw=blue!80,line width=0.3mm},   
    pretrain_work/.style =           {align=center, text width=8.5cm,rounded corners=3pt, fill=blue!10,draw=blue!0,line width=0.3mm},  
    template/.style =           {align=center,text width=2.0cm,rounded corners=3pt, fill=red!10,draw=red!80,line width=0.3mm},   
    template_work/.style =           {align=center,text width=8.5cm,rounded corners=3pt, fill=red!10,draw=red!0,line width=0.3mm},    
    answer/.style =           {align=center,text width=2.0cm,rounded corners=3pt, fill= cyan!10,draw= cyan!80,line width=0.3mm},   
    answer_work/.style =           {align=center,text width=8.5cm,rounded corners=3pt, fill= cyan!10,draw= cyan!0,line width=0.3mm},      
    multiple/.style =           {align=center,text width=2.0cm,rounded corners=3pt, fill= orange!10,draw= orange!80,line width=0.3mm},   
    multiple_work/.style =           {align=center,text width=8.5cm,rounded corners=3pt, fill= orange!10,draw= orange!0,line width=0.3mm},        
    tuning/.style =           {align=center,text width=2.0cm,rounded corners=3pt, fill= magenta!10,draw= magenta!80,line width=0.3mm},   
    tuning_work/.style =           {align=center,text width=8.5cm,rounded corners=3pt, fill= magenta!10,draw= magenta!0,line width=0.3mm},          
}

\newcommand{\lstbg}[3][0pt]{{\fboxsep#1\colorbox{#2}{\strut #3}}}

\lstdefinelanguage{diff}{
  basicstyle=\ttfamily\small,
  morecomment=[f][\lstbg{red!20}]-,
  morecomment=[f][\lstbg{green!20}]+,
}

\lstdefinelanguage{diffpython}{
  language=diff,
  morekeywords={def, if, else, for, while, return, import, from, as, class, with, try, except, finally, raise, lambda, and, or, not, in, is, None, True, False},
  morecomment=[l]{\#},
  morestring=[b]",
  morestring=[b]',
}

\title{\texttt{FlowPIE}: Test-Time Scientific Idea Evolution with Flow-Guided Literature Exploration}

\author{
    \scriptsize 
    Qiyao Wang$^{1,2,*}$\quad 
    Hongbo Wang$^{3,*}$\quad
    Longze Chen$^{1,2}$\quad
    Zhihao Yang$^{1,2}$\quad
    Guhong Chen$^{1}$\quad
    Hamid Alinejad-Rokny$^{4}$\quad
    Hui Li$^{6}$\quad
    Yuan Lin$^{3\dagger}$\quad
    Min Yang$^{1,5\dagger}$ \\
    $^1$~\raisebox{-0.3ex}{\includegraphics[height=2ex]{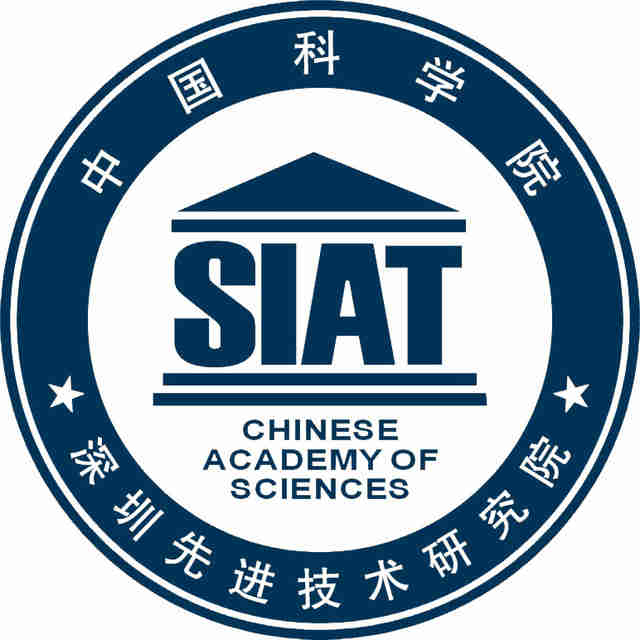}}~Shenzhen Institute of Advanced Technology, Chinese Academy of Sciences\quad
    $^2$~\raisebox{-0.2ex}{\includegraphics[height=2ex]{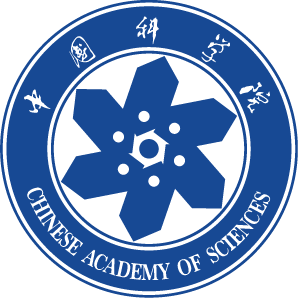}}~University of Chinese Academy of Sciences\\
    $^3$~\raisebox{-0.2ex}{\includegraphics[height=2ex]{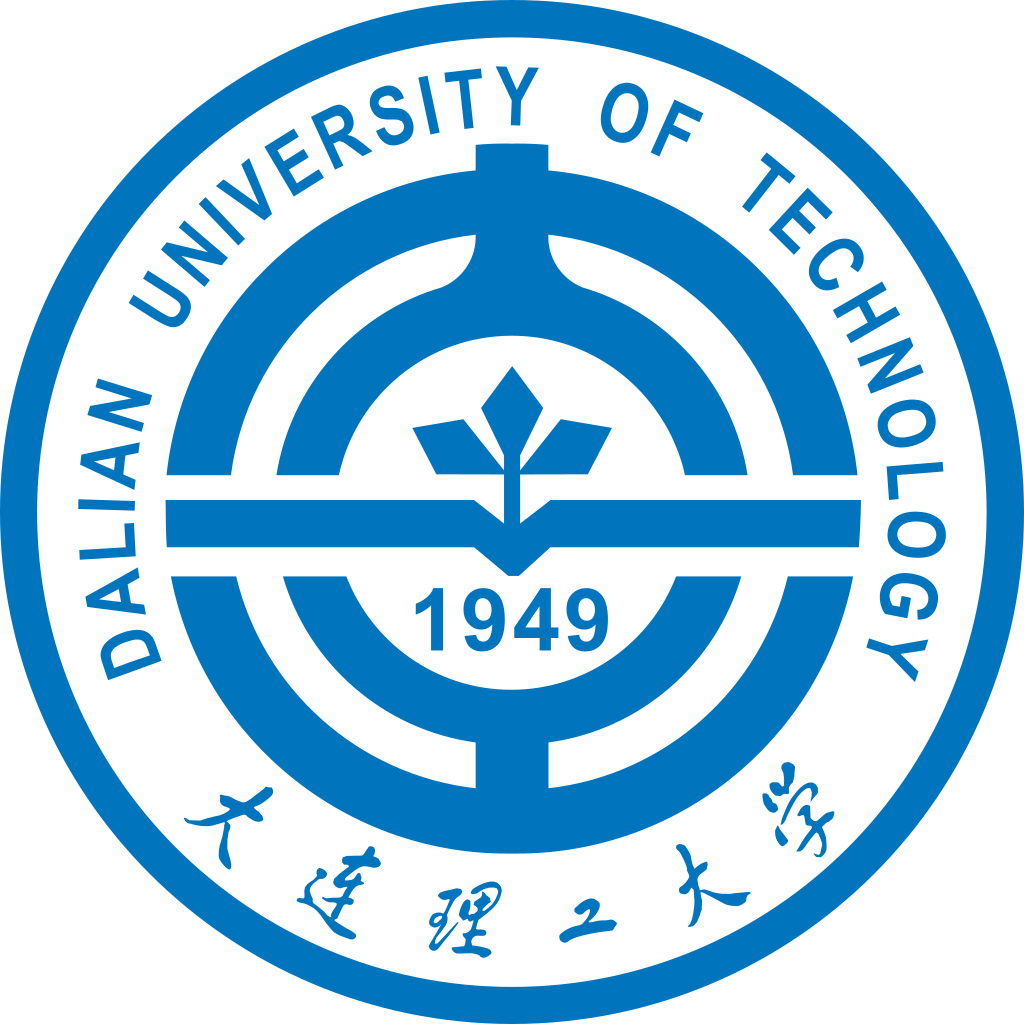}}~Dalian University of Technology\text{ }
    $^4$~\raisebox{-0.2ex}{\includegraphics[height=2ex]{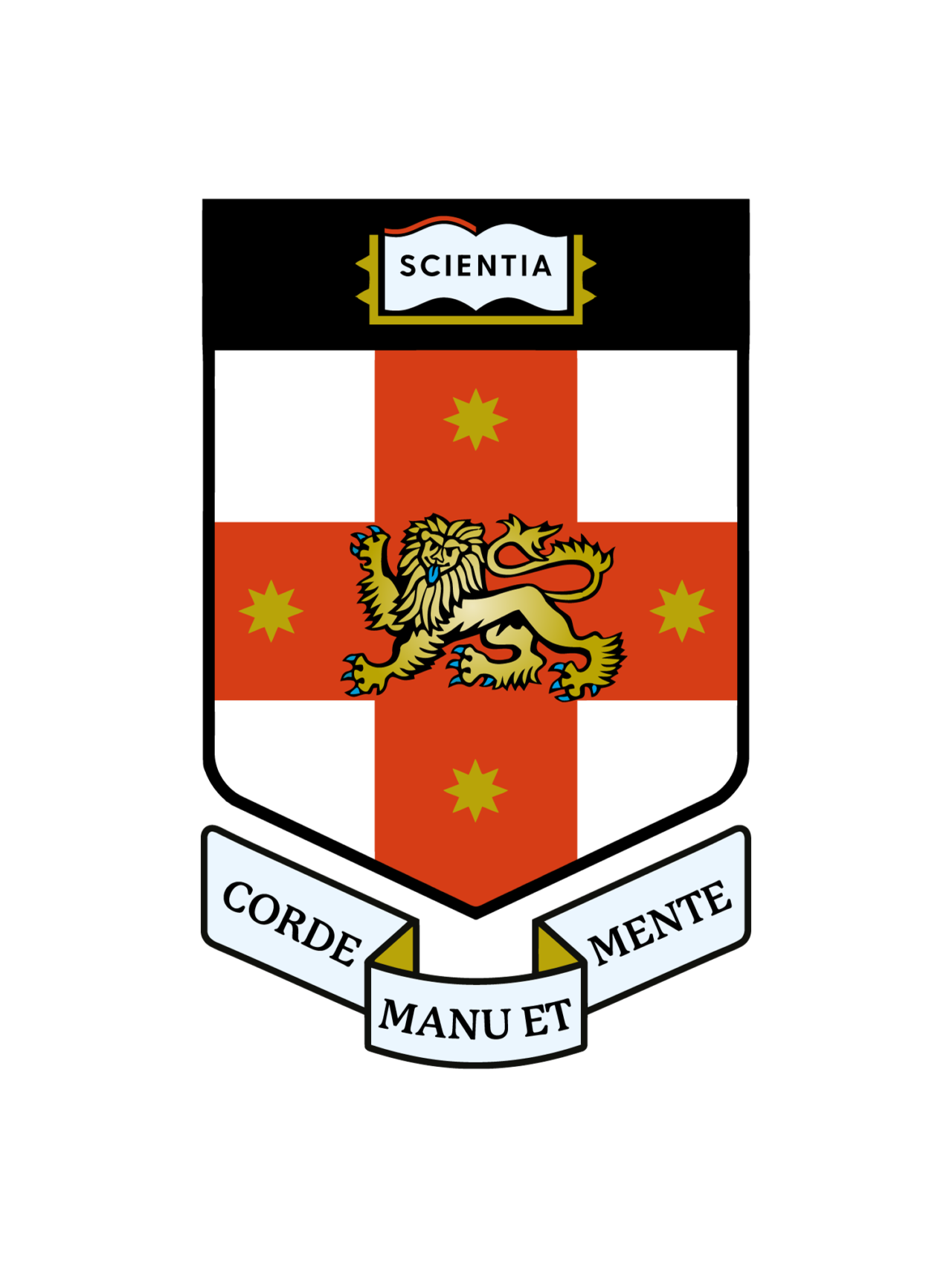}}~UNSW Sydney\text{ }
    $^5$~\raisebox{-0.2ex}{\includegraphics[height=2ex]{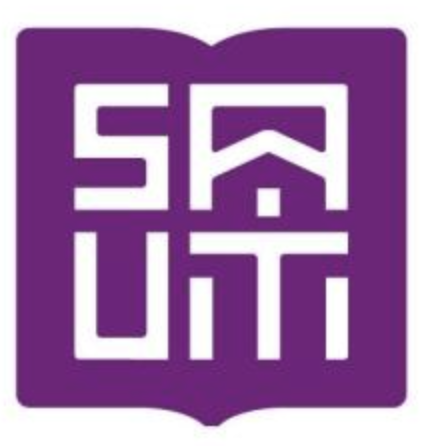}}~Shenzhen University of Advanced Technology\text{ }
    $^6$~\raisebox{-0.3ex}{\includegraphics[height=2ex]{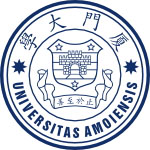}}~Xiamen University\\
    \faEnvelope[regular]~\texttt{wangqiyao25@mails.ucas.ac.cn}  \quad
    \faEnvelope[regular]~\texttt{zhlin@dlut.edu.cn}  \quad
    \faEnvelope[regular]~\texttt{min.yang@siat.ac.cn}  \\
    \faHome~\href{https://flowpie.wangqiyao.me/}{Website} \quad \faGithub~\href{https://github.com/AIforIP/FlowPIE}{FlowPIE} \quad
    $^*$ Equal Contribution. \quad
    $^\dagger$ Corresponding Authors. 
}
\begin{document}

\begin{abstract}
Scientific idea generation (SIG) is critical to AI-driven autonomous research, yet existing approaches are often constrained by a static retrieval-then-generation paradigm, leading to homogeneous and insufficiently divergent ideas. In this work, we propose \textbf{FlowPIE}, a tightly coupled retrieval–generation framework that treats literature exploration and idea generation as a co-evolving process. FlowPIE expands literature trajectories via a flow-guided Monte Carlo Tree Search (MCTS) inspired by GFlowNets, using the quality of current ideas assessed by an LLM-based generative reward model (GRM) as a supervised signal to guide adaptive retrieval and construct a diverse, high-quality initial population. Based on this population, FlowPIE models idea generation as a \textbf{test-time idea evolution process}, applying selection, crossover, and mutation with the isolation island paradigm and GRM-based fitness computation to incorporate cross-domain knowledge. It effectively mitigates the information cocoons arising from over-reliance on parametric knowledge and static literature. Extensive evaluations demonstrate that FlowPIE consistently produces ideas with higher novelty, feasibility and diversity compared to strong LLM-based and agent-based frameworks, while enabling reward scaling during test time. 
\end{abstract}
\maketitle 

\section{Introduction}

With the rapid development of large language models (LLMs)~\citep{hurst2024gpt,liu2024deepseek}, their strong multidisciplinary understanding and reasoning capabilities make it increasingly feasible to synthesize knowledge from large-scale scientific literature. Recent efforts have shown that LLM-based and agent-based systems can support the entire scientific research pipeline, from proposal development and experimental design to result analysis and paper drafting, forming an autonomous research paradigm~\citep{2025arXiv250701903C}.

\begin{wrapfigure}{r}{0.54\linewidth}
\includegraphics[width=\linewidth]{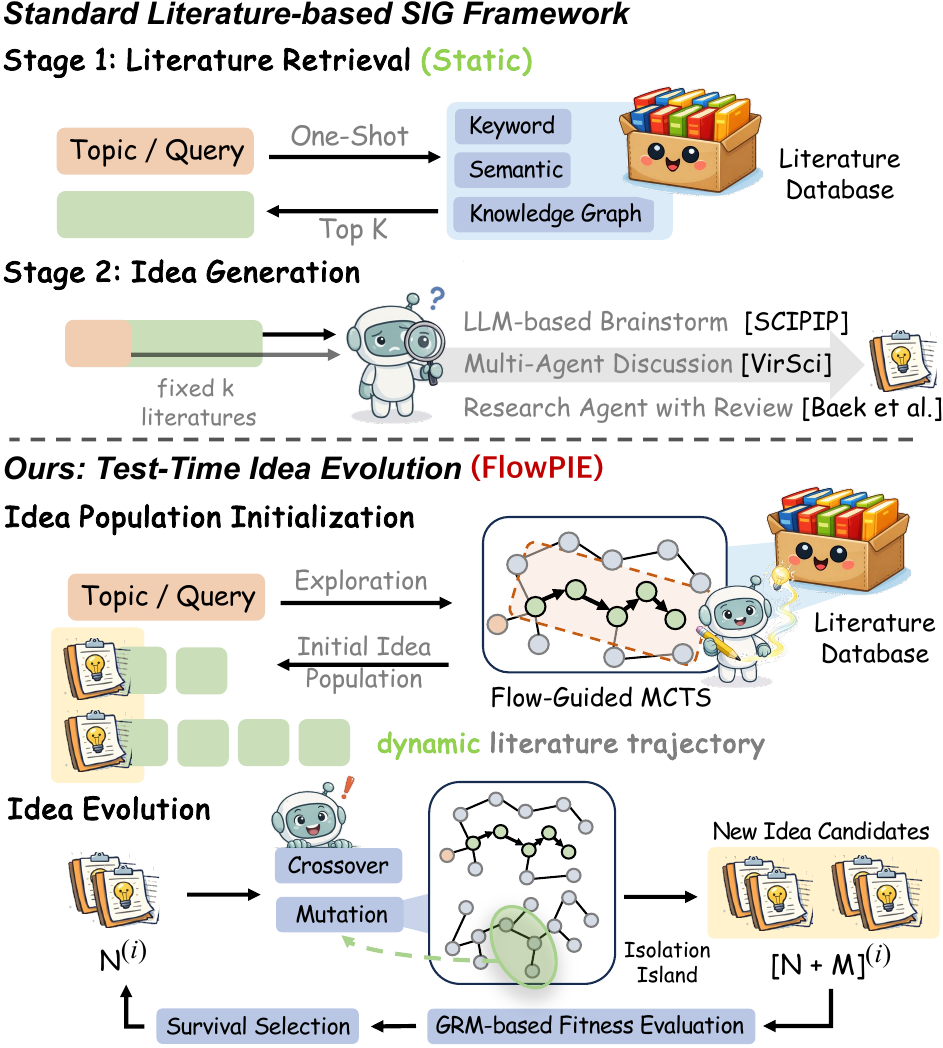}
    \caption{{Comparison of traditional literature-based SIG frameworks and our FlowPIE.}}
    \label{fig:comparison}
    \vspace{-1.5em}
\end{wrapfigure}

Scientific idea generation (SIG) has emerged as a key frontier in autonomous research, attracting significant efforts across diverse domains. As shown in Figure~\ref{fig:comparison}, most existing methods mine novel ideas from literature databases using a decoupled two-stage framework: first \textit{retrieving relevant literature}, and then \textit{generating ideas based on the retrieved literature}. This pipeline typically relies on a single retrieval step driven by keyword matching and semantic relevance~\citep{wang2024scipip} to a specific topic. However, relying on this static manner as the sole source of the inspiration yields contexts that are merely topically similar, rather than genuinely conducive to innovation. Consequently, this restricts the depth and breadth of the provided knowledge, frequently leading to homogeneous ideas with limited divergence.

In the idea generation stage, prior works leverage LLMs brainstorming~\citep{wang2024scipip}, research agent with review~\citep{baek2025researchagent} or multi-agent discussion~\citep{su2025many} to generate and refine
ideas. These approaches attempt to exploit the parametric knowledge encoded in LLMs together with information from static retrieval literature. However, such designs risk trapping LLM-based generators within an information cocoon, bounded by their internal knowledge and static 
external sources. These limitations motivate us to revisit the widely adopted retrieval-and-generation paradigm for SIG. Specifically, we focus on the following two research questions (RQs): \textbf{\textit{RQ1}}: How can literature retrieval be a \textit{dynamic}, \textit{adaptive} component within the idea generation, instead of a \textit{static} stage? \textbf{\textit{RQ2}}: How can LLMs leverage retrieved literature and their relationships to generate novel and divergent ideas and continuous refinement?

In this work, we propose \textbf{FlowPIE}, a tightly coupled retrieval–generation framework for test-time idea evolution, as illustrated in Figure~\ref{fig:flowpie-framework}. 
Moving beyond the traditional static retrieval paradigm, FlowPIE unifies literature retrieval and idea generation into a dynamic and adaptive \textbf{\textit{test-time idea evolution process}}. Within this process, intermediate generated ideas serve as active feedback to guide subsequent literature exploration. Specifically, it performs structured exploration over the literature subgraph by modeling the retrieval as a flow-guided MCTS inspired by GFlowNets~\citep{Bengio2021GFlowNetF}, thereby incrementally expanding retrieval trajectories in both breadth and depth. 

The ideas generated along these paths are then organized into an initial population for subsequent iterative evolution. FlowPIE is implemented as an evolutionary algorithm that operates over this idea population through the iterative application of selection, crossover, and mutation operators, with fitness evaluated via LLM-based generative reward model~(GRM). During the mutation stage of the evolutionary process, FlowPIE specifically introduces the \textit{isolation island} paradigm. This paradigm maintains multiple isolated literature, thereby facilitating the incorporation of cross-domain knowledge and diverse literature characteristics. 

\begin{wrapfigure}{r}{0.55\linewidth}
    \vspace{-1.2em}
\includegraphics[width=0.9\linewidth]{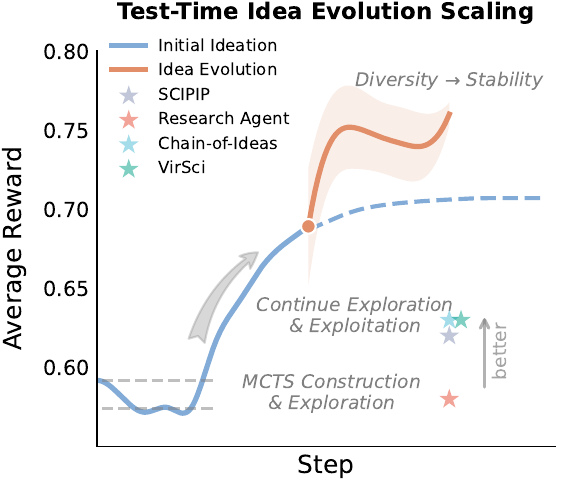}
    \vspace{-0.5em}
    \caption{{Test-time idea evolution scaling for reward.} \textbf{\textcolor[HTML]{82ACD7}{Initial Ideation}} couples literature exploration with idea generation via flow-guided MCTS, where higher reward, reflecting better initial ideas, amplify the weight of corresponding literature. \textbf{\textcolor[HTML]{E18E69}{Idea Evolution}} uses various evolutionary operators to guide ideas toward regions of higher rewards and stable convergence.}
    \label{fig:reward-on-ai-idea-bench}
    \vspace{-1em}
\end{wrapfigure}

Extensive experiments and human evaluations on AI Idea Bench 2025~\citep{qiu2025aiideabench2025} and IdeaBench~\citep{guo2025ideabench} demonstrate that FlowPIE outperforms prior LLM-based and agent-based baselines, while generating more novel, divergent ideas and demonstrates domain generalization.
Beyond benchmark results, we analyze the reward scaling curve of FlowPIE in Figure~\ref{fig:reward-on-ai-idea-bench}, observing that its reward initially fluctuates during literature exploration and then rises due to flow-guided balancing of exploration and exploitation. After obtaining initial ideas, continuous idea evolution further refines these ideas toward regions of higher quality and more stable convergence. Notably, both the evolved ideas and even the initial ideas achieve higher reward scores than those of other baselines.
\textbf{Our main contributions are as follows:}
\begin{itemize}
    \item We propose the novel framework, \textbf{FlowPIE}, which models idea generation as a test-time idea evolution process, iteratively applying survival selection, crossover, and mutation operators to an initial population of ideas, supervised by a GRM-based fitness evaluation.
    \item We rethink the retrieval-generation SIG framework and propose a novel \textbf{flow-guided MCTS} in FlowPIE that integrates dynamic literature retrieval with initial idea generation, leveraging idea quality feedback to balance exploration and exploitation in literature retrieval.
    \item Experimental results on benchmarks demonstrate that our FlowPIE significantly improves idea quality and exhibits domain generalization. Notably, analysis of the idea evolution reward curve shows that FlowPIE exhibits clear test-time scaling on reward and consistently surpasses other baselines.
\end{itemize}

\section{Related Work}
\paragraph{AI for Science Research.} With the advancement of LLMs, the landscape of scientific inquiry has been fundamentally reshaped across a wide range of disciplines, including physics~\citep{ye2025physics}, medicine~\citep{liao-etal-2024-medcare}, and mathematics~\citep{RomeraParedes2023MathematicalDF}. \citet{2025arXiv250701903C} established a holistic framework of AI for Research that not only systemizes current AI applications but also explores the future trajectory of AI's impact on the research ecosystem. AI-Scientist~\citep{lu2024ai} and AI-Researcher~\citep{tang2025airesearcherautonomousscientificinnovation} aim to support the complete lifecycle of autonomous research, with the ultimate goal of producing a full research paper, including code generation and experimental execution.

\paragraph{Scientific Idea Generation.} Most previous SIG algorithms are rooted in simulations of the human ideation process. SCIPIP~\citep{wang2024scipip} leverages keywords and semantic similarity to statically retrieve relevant literature and synthesizes new ideas through LLM-based brainstorming. 
\citet{li2024chain} consider relationships among prior literature and leverage a CoI agent to construct a chain-of-ideas before idea generation, using it as a curated future direction prompt for subsequent synthesis. 
\citet{baek2025researchagent} construct an entity-centric knowledge graph for literature survey, then generate and review ideas using a research agent and a review agent.
Additionally, VirSci~\citep{su2025many} adopts a multi-agent system, including team construction and discussion for ideation simulation. 

We concur with these literature-based methods that new ideas arise from prior art rather than in a vacuum, and further argue that the quality of generated ideas is constrained by the quality of the relevant prior works. Therefore, our proposed FlowPIE approaches idea generation from a test-time idea evolution perspective based on an evolutionary algorithms (EAs), and couples literature retrieval with the quality of the initial idea to enable high-quality literature trajectory exploration. More discussion about related work on the evaluation of SIG and LLMs-enhanced EAs framework are provided in Appendix~\ref{sec:more-related-work}.

\section{Methodology}
\begin{figure*}[!t]
\includegraphics[width=1\linewidth]{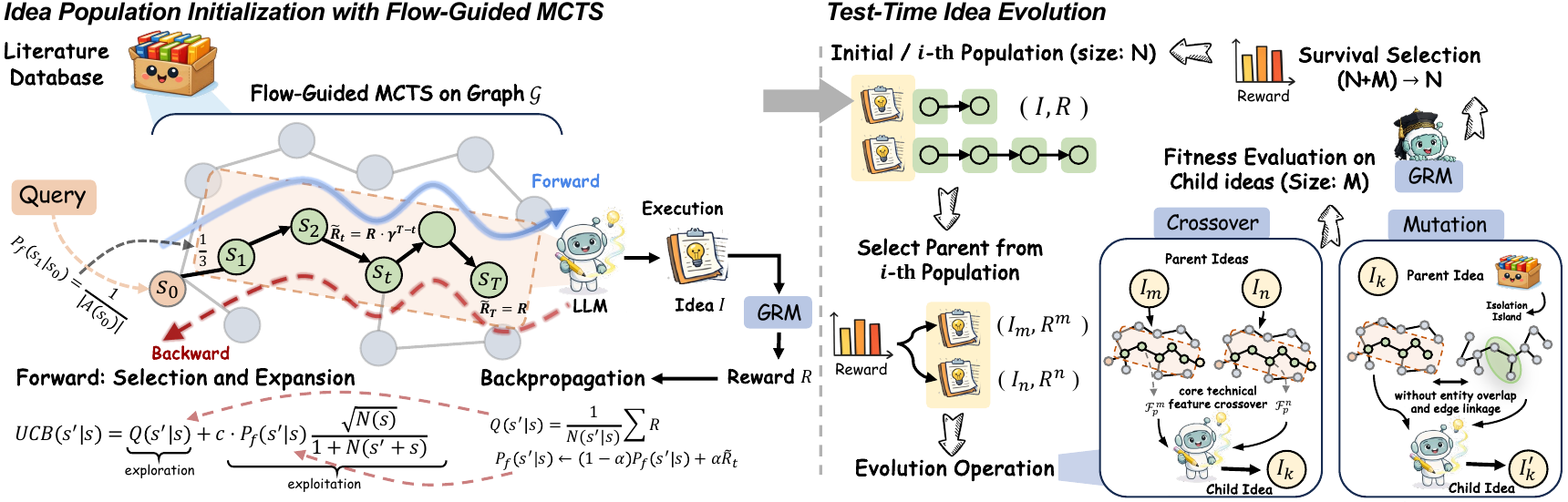}
    \caption{Overview of FlowPIE. \textbf{Left:} Idea initialization based on flow-guided MCTS: forward exploration via flow-guided UCB~(Eq.\ref{UCB}) for nodes selection and expansion, and backward updating~(Eq.\ref{flow}) to enforce local and global flow constraints; this dual process dynamically adjusts literature weights with GRM-based rewards on generated ideas. \textbf{Right:} Test-time idea evolution, supervised by GRM-based fitness computation, leveraging various operators, including crossover on core technical features and isolation-island-enhanced mutation, to evolve ideas.}
    \label{fig:flowpie-framework}
\end{figure*}

In this paper, we contend that the generation of scientific ideas is not an isolated process. 
Rather, their novelty must be grounded in technical realism, emerging cumulatively from the synthesis of prior sciences knowledge. 
Based on this perspective, we propose \textbf{FlowPIE}, as illustrated in Fig.~\ref{fig:flowpie-framework} and Algorithm~\ref{algorithm-1}, and~\ref{algorithm-2}. 
It models the SIG process using an evolutionary algorithm. 
Specifically, the initial idea population is generated through dynamic literature exploration drawn by a flow-guided Monte Carlo Tree Search (MCTS) (see Sec.~\ref{initial-ideation}).
The initial population is then refined through iterative evolution incorporating fitness evaluation, survival selection, and crossover and mutation operators, as detailed in Sec.~\ref{idea-evolution}.

\paragraph{Task and Idea Formulation.}
The target of scientific idea generation is to formulate novel and diverse research hypotheses that can accelerate automated scientific discovery. Given a topic or query $q$, idea generator aims to generate a set of structured scientific ideas grounded in existing knowledge and literature. Formally, the framework leverages an LLM-based idea generator to map a $q$ to an idea set $\mathcal{I}=\{I_1,...I_k\}$, where each idea $I$ is represented as a structured tuple comprising \textbf{\textit{Motivation}}, \textbf{\textit{Method}} and \textbf{\textit{Experimental Plan}}. 
Prior work such as SCIPIP, which primarily focuses on problem–method, our formulation explicitly incorporates a detailed experimental plan, following \citet{baek2025researchagent}, making the ideas more actionable and better aligned with practical research workflows. 

\subsection{Initial Idea Population Construction}
\label{initial-ideation}

\paragraph{Patent Literature Graph Construction.}
Most prior works leverage papers as the literature source. In contrast, we use patents accessing from the USPTO, whose clearly defined, precisely scoped and easily extracted structural claims reduce ambiguity in scientific statements, thereby enabling more stable and reliable generation. 
To model the relationships among literature, we construct a hierarchical structured attribution for each patent. Consider the literature database $\mathcal{D}$ spanning various domains within the International Patent Classification~(IPC), given any patent $p \in \mathcal{D}$, we leverage an LLM maps $p$ into an attribution tuple $\langle \mathcal{A}_p, \mathcal{F}_p, \mathcal{S}_p \rangle$, where $\mathcal{A}_p$ represents its abstract, $\mathcal{F}_p$ represents its core technical feature set extracted by a LLM, and $\mathcal{S}_p \in \mathbb{R}^d$ denotes its semantic embedding.

We formalize the whole literature graph as $\mathcal{G}=(\mathcal{V,E})$, where $\mathcal{V}$ is the node set of unique patent entities parsed from $\mathcal{D}$. 
The edge set $\mathcal{E}$ represents the relation between nodes, where an edge $e_{ij} \in \mathcal{E}$ exists if patents $p_i$ and $p_j$ satisfy at least one of the following criteria: \textbf{(\textit{i})} a direct citation relation; \textbf{(\textit{ii})} at least an overlap in core technical features, \textit{i.e.} $\mathcal{F}_i\cap\mathcal{F}_j\ne \varnothing$; \textbf{(\textit{iii})} the semantic similarity $sim(\mathcal{E}_i,\mathcal{E}_j)$ exceeding the threshold. 
Details of patent literature are in Appendix~\ref{appendix:patent-literature}.

\paragraph{Idea Initialization with Literature Exploration via Flow-Guided MCTS.}
To construct a high-quality and diverse initial population of ideas, inspired by GFlowNet~\citep{Bengio2021GFlowNetF}, we propose a literature exploration mechanism, termed flow-guided MCTS, over graph $\mathcal{G}$.
Given a query, we regard it as the root node $s_0$ and set its initial flow $F(s_0)=1$, then we retrieve relevant literature using similarity. For any node $s$  with expandable adjacent node $s'\in A(s)$, the flow $P_{f}$ is uniformly initialized as $P_{f}(s'\mid s) = \frac{1}{|A(s)|}$. \\
\noindent\textbf{Selection and Expansion:} We balance exploration and exploitation when selecting and expanding new adjacent nodes along the edges of the constructed literature graph $\mathcal{G}$, by utilizing a flow-guided Upper Confidence Bound (UCB):
\begin{equation}
    UCB(s'\mid s) = Q(s'\mid s) + c \cdot P_{f}(s'\mid s) \frac{\sqrt{N(s)}}{1 + N(s'\mid s)}
    \label{UCB}
\end{equation}
where $Q(s'\mid s)$ denotes the expected value, encouraging exploitation of paths that have previously generated high reward ideas.  $N(\cdot)$ denotes visit counts, and $c$ is the exploration rate. 
\noindent\textbf{Execution and Backpropagation:} The LLM-based idea generator produces an idea based on the currently explored patent trajectory, which then receives a reward $R$ from the GRM. We then backpropagate this reward to update the UCB value of trajectory in Equation~\ref{UCB}. Considering the importance of literature at different depths, we introduce a depth-decayed reward $\tilde{R}_t = R  \cdot \gamma ^{T-t}$, where $T$ is the maximum depth of current trajectory. The value estimation is updated via standard averaging $Q(s'\mid s) = \frac{1}{N(s'\mid s)}\sum R^i$, while the flow probability is updated using a moving average:
\begin{equation}
    P_{f}(s'\mid s)\leftarrow(1-\alpha)P_{f}(s'\mid s) + \alpha \tilde{R}_t
    \label{flow}
\end{equation}
where $\alpha \in [0,1]$ controls the weight of the reward.

Following this, $P_{f}$ is locally normalized over $A(s_t)$ at each time step $t$. Crucially, $P_{f}$ acts as a local probability constrained by the global flow $F$, defined as $P_{f}(s'\mid s)=\frac{F(s'\mid s)}{F(s)}$. Thus, the global flow iteratively updates forwardly via $F(s_{i+1}) = F(s_i)\cdot P_{f}(s_{i+1}\mid s_i)$. 
The iterative exploration terminates once the reward variance of the generated ideas falls below a threshold \(\epsilon\). These ideas subsequently serve as the initial population for the idea evolution phase, accompanied by the explored literature traced from the root node $s_0$. 

\subsection{Test-Time Idea Evolution}
\label{idea-evolution}
The initial idea population within the flow-guided MCTS primarily serves as an intermediate signal for broad and deep literature exploration, but lacks sufficient continuous refinement to enhance novelty and feasibility, resulting in the reward bottleneck shown in Figure~\ref{fig:reward-on-ai-idea-bench}. In this section, we introduce test-time idea evolution within FlowPIE, iteratively applying survival selection, crossover, and mutation operators for LLM-based idea generator to the initial population for continuous evolution. 

\paragraph{Idea Evolution with Crossover and Mutation.}
We leverage an LLM-based idea generator, to produce offspring ideas through pairwise crossover and isolation-island-enhanced mutation operator.\\
\noindent\textbf{Crossover Operator.}
The crossover operator aims to synthesize advantageous from different promising ideas. Given two parent ideas $\mathcal{I}_m$ and $\mathcal{I}_n$, along with the explored literature, we define the crossover process as $\mathcal{I}_k=\text{LLM}(q,\mathcal{I}_m,\mathcal{I}_n)$, where $\mathcal{I}_k$ denotes the generated offspring idea. 
Rather than performing superficial textual interpolation, the operator recombines the core technical features of the two parent ideas under the guidance of the retrieved literature, enabling the LLM to synthesize a novel descendant idea that integrates their complementary characteristics and inherits their strengths.\\
\noindent\textbf{Mutation Operator with Isolation Island.}
To prevent idea evolution from being trapped in local optima while maintaining diversity, we introduce a mutation operator governed by a mutation rate $\rho$. 
For each offspring $\mathcal{I}_k$, mutation is triggered by sampling $z_k\sim \text{Bernoulli}(\rho)$. If $z_k=1$, we apply the \textbf{\textit{lsolation Island}} strategy on graph $\mathcal{G}$.
Instead of retrieving literature only from the current local neighborhoods, we sample an auxiliary set $\mathcal{Q}_k^{\text{iso}}\subset \mathcal{G}$ from topologically distant subgraphs disconnected from the current neighborhood.
The mutated idea is then generated by $\mathcal{I}'_k=\text{LLM}(q,\mathcal{I}_k,\mathcal{Q}_k^{\text{iso}})$, where the LLM is encouraged to logically integrate the out-of-domain (OOD) information into $\mathcal{I}_k$ which enrich the boundaries of the ideas.

\paragraph{Idea Fitness Evaluation.}
The offspring ideas obtained through evolution are then evaluated using the GRM. Specifically, we use the GRM to assess each idea across multiple dimensions (e.g., novelty and feasibility), which are subsequently aggregated into a scalar fitness score.
Reward definitions and prompts detials are provided in Appx.~\ref{appendix:grm-metric} and~\ref{appendix:prompts}.

\paragraph{Survival Selection.}
We adopt a tournament selection strategy to form the next-generation population $\mathcal{I}_{next}$. Specifically, the offspring and parent ideas are merged into a candidate pool $C$. 
While $|\mathcal{I}_{next}|<N$, we randomly sample a subset $S\subset C$, select the highest-reward idea, add it to $\mathcal{I}_{next}$. The process repeats until $|\mathcal{I}_{next}|=N$, thereby preserving high-fitness ideas for the next generation. The evolution process stops when the maximum number of iterations is reached or the reward converges, yielding the final $N$ evolved ideas.

\section{Experiments}

\subsection{Experimental Setup}
\label{sec:4.1}

\noindent\textbf{Baselines.} We compare FlowPIE with two types of baselines, including \textbf{(\textit{i}) LLM-based Framework}, SCIPIP~\citep{wang2024scipip}, which leverages a dual-path framework that integrates retrieved literature with LLM-based brainstorming; \textbf{(\textit{ii}) Agent-based Framework}, Research Agent~\citep{baek2025researchagent} iteratively leverages a research agent and a review agent, while Chain-of-Ideas~\citep{li2024chain} employs a CoI-Agent to model dependency relations among prior works. We also compare against the multi-agent baseline VirSci~\citep{su2025many}, which enables discovery through simulated team construction and structured discussion. To ensure a fair comparison, we use \texttt{GPT-4o-mini} as the idea generator for all methods. 
Implementation details and method costs are provided in Appendix~\ref{appendix:implementation-details}.

\paragraph{Evaluation Benchmarks and Metrics.} We evaluate all methods on two SIG benchmarks. \textbf{(\textit{i}) AI Idea Bench 2025}~\citep{qiu2025aiideabench2025}, which comprises papers from top AI conferences such as ICLR, CVPR and ACL, as the target idea source. It contains three main tasks: idea-to-topic matching (I2T), idea-to-idea matching (I2I), and idea multiple-choice evaluation (IMCQ). The first two tasks use an LLM-as-a-judge paradigm with scores ranging from 1 to 5, while IMCQ uses accuracy. \textbf{(\textit{ii})  IdeaBench}~\citep{guo2025ideabench}, which contains 2,374 influential biomedical papers, evaluates generated ideas using two similarity‑based metrics: BERTScore for semantic similarity, with a practical upper limit of 0.718 reported in the original paper and idea overlap on a 0-10 scale; it also uses two insight score for novelty and feasibility, computed from the relative ranking of generated ideas against the target paper’s idea. 
For fair and consistent evaluation, all metrics are assessed using the frontier \texttt{GPT-5-mini} model. Detailed task and metric definitions are provided in Appendix~\ref{appendix:benchmark-metrics}. 

\paragraph{Human Evaluation Setup.}We follow the criteria of~\citet{sican}, employing human experts who are computer science PhD students to blindly evaluate 20\% randomly sampled ideas from AI Idea Bench 2025 per method on novelty, feasibility, excitement, and expected effectiveness using a 10-point scale. Details are provided in Appendix~\ref{appendix:human-evaluation}.

\subsection{Results}
\begin{table*}[!h]
 
\setlength{\tabcolsep}{4.8mm}
    \scalebox{0.77}{\begin{tabular}{lccccccc}
    \toprule
    \multirow{2}[4]{*}{\textbf{Method}} & \multicolumn{2}{c}{\textbf{I2T}} & \multicolumn{2}{c}{\textbf{I2I}} & \multicolumn{2}{c}{\textbf{IMCQ}} \\
\cmidrule{2-7}         & \multicolumn{1}{l}{\textbf{Motivation}} & \multicolumn{1}{l}{\textbf{Exp Plan}} & \multicolumn{1}{l}{\textbf{Motivation}} & \multicolumn{1}{l}{\textbf{Exp Plan}} & \multicolumn{1}{l}{\textbf{Motivation}} & \multicolumn{1}{l}{\textbf{Exp Plan}} \\
    \midrule
    \multicolumn{1}{l}{\textbf{SCIPIP}}        &  4.18$\pm$0.662     &   —    &    3.68$\pm$0.383   & —      &  0.464     &—  \\
    \multicolumn{1}{l}{\textbf{Research Agent}}        &   4.56$\pm$0.679    & 3.59$\pm$0.670      &   3.78$\pm$0.491   &   3.45$\pm$0.619    &  0.510     & 0.497 \\
    \multicolumn{1}{l}{\textbf{Chain-of-Ideas}}        &  \underline{4.63$\pm$0.703}     & \textbf{3.79$\pm$0.566}      &  3.74$\pm$0.504     & 3.50$\pm$0.653     &  0.507     & 0.433 \\
    \multicolumn{1}{l}{\textbf{VirSci}}        &   4.32$\pm$0.652    &  3.68$\pm$0.565     & 3.95$\pm$0.410      & \underline{3.84$\pm$0.410}      &   0.560    & 0.580 \\
    \rowcolor{gray!25} \multicolumn{1}{l}{\textbf{FlowPIE~(\textit{Ours})}}        & \textbf{4.64$\pm$0.606}      & \underline{3.72$\pm$0.481}      & \textbf{4.44$\pm$0.318}      & \textbf{3.85$\pm$0.347}      & \textbf{0.780}      & \textbf{0.635} \\
    \rowcolor{gray!15}\quad\textbf{Initial Population}        &   4.58$\pm$0.616    &  3.58$\pm$0.512     &  \underline{4.26$\pm$0.224}     &  3.61$\pm$0.235     &  \underline{0.753}     & \underline{0.622} \\
    \bottomrule
    \end{tabular}}
    \caption{{Main results on AI-Idea-Bench-2025}. We generate three ideas with each method and report using the \textit{mean $\pm$ std} for I2T and I2I. \textbf{Bold} indicates the best score, and \underline{underlining} denotes the second-best score.}
  \label{tab:main-results-ai-idea-bench}%
\end{table*}%
\begin{wraptable}{r}{0.5\textwidth}
\vspace{-1em}
 
\setlength{\tabcolsep}{2.5mm}
\scalebox{0.8}{
\begin{tabular}{lcccc}
\toprule
\textbf{Method}&\textbf{Simi.}&\textbf{Overlap}&\textbf{NI} &\textbf{FI} \\
\midrule
\textbf{SCIPIP} & 0.526 & 5.03 & 0.816 & 0.133 \\
\textbf{Research Agent} & \underline{0.558} & 6.66 & 0.722 & \textbf{0.138} \\
\textbf{Chain-of-Ideas} & 0.482 & 7.24 & \textbf{0.926} & 0.095 \\
\textbf{VirSci} & 0.521 & 6.24 & 0.716 & 0.075 \\
\rowcolor{gray!25}\textbf{FlowPIE(\textit{Ours})} & \textbf{0.559} & \textbf{7.76} & \underline{0.825} & 0.105 \\
\rowcolor{gray!15}\textbf{Initial Population} & 0.532 & \underline{7.64} & 0.750 & \underline{0.136} \\
\bottomrule
\end{tabular}}
\vspace{-0.5em}
\caption{IdeaBench Results. \textit{Simi.} denotes Semantic Similarity, while \textit{Overlap} denotes Idea Overlap. \textit{NI} and \textit{FI} denote the Novelty and Feasibility Insight scores.}
\label{tab:ideabench}
\vspace{-1.5em}
\end{wraptable}
\noindent\textbf{Benchmark Results.} As shown in Table~\ref{tab:main-results-ai-idea-bench} and Table~\ref{tab:ideabench}, we report results on AI Idea Bench 2025 and IdeaBench, respectively. Across the three tasks of AI Idea Bench 2025, FlowPIE generates ideas that demonstrate high consistency with the target topic and strong relevance to the idea of target paper, and is the only method to obtain a motivation score above 4 in I2I task. Compared with competing candidate ideas, it achieves a motivation selection accuracy of 0.780 and experiment plan selection accuracy of 0.635 in the IMCQ task when selecting the best idea among alternatives. 

For IdeaBench, our FlowPIE achieves the highest Semantic Similarity and Idea Overlap with the target paper. Considering the Novelty Insight Score (NI) and Feasibility Insight Score (FI), FlowPIE and its initial population both lie on the Pareto front, indicating a competitive balance between novelty and feasibility. In particular, FlowPIE achieves a well-balanced trade-off between NI and FI.

\begin{wraptable}{r}{0.5\textwidth} 
\setlength{\tabcolsep}{1mm}
\scalebox{0.73}{
\begin{tabular}{lccc|ccccc}
\toprule
\multirow{2}{*}{\textbf{Method}}
& \multicolumn{3}{c}{\textbf{Reward}} 
& \multicolumn{5}{c}{\textbf{Human Evaluation}} \\
\cmidrule(lr){2-4} \cmidrule(lr){5-9}
& \textbf{N} & \textbf{F} & \textbf{Avg.} 
& \textbf{N} & \textbf{F} & \textbf{E} & \textbf{EE} & \textbf{Avg.} \\
\midrule
\textbf{SCIPIP} & 0.50 & \underline{0.73} & 0.62 & \underline{0.38} & \textbf{0.37} & 0.30 & \underline{0.29} & \underline{0.34} \\
\textbf{Research Agent} & 0.49 & 0.66 & 0.58 & 0.22 & 0.19 & 0.13 & 0.13 & 0.17 \\
\textbf{Chain-of-Ideas} & 0.59 & 0.66 & 0.63 & 0.34 & 0.29 & \underline{0.31} & \underline{0.29} & 0.31 \\
\textbf{VirSci} & 0.60 & 0.65 & 0.63 & 0.27 & 0.17 & 0.20 & 0.20 & 0.21 \\
\rowcolor{gray!25}\textbf{FlowPIE (\textit{Ours})} & \textbf{0.75} & \textbf{0.76} & \textbf{0.76} & \textbf{0.45} & \underline{0.36} & \textbf{0.38} & \textbf{0.37} & \textbf{0.39} \\
\rowcolor{gray!15}Initial Population & 0.68 & 0.68 & 0.68 & -- & -- & -- & -- & -- \\
\rowcolor{gray!5}\textit{w/o} isolation island & \underline{0.72} & \underline{0.73} & \underline{0.73} & -- & -- & -- & -- & -- \\
\bottomrule
\end{tabular}}
\caption{Reward and human evaluation results. N: Novelty, F: Feasibility, E: Excitement, EE: Expected Effectiveness, and Avg.: Average.}
\label{tab:reward-results}
\vspace{-1em}
\end{wraptable}
Overall, FlowPIE achieves superior performance on both benchmarks and lies on the Pareto frontier. Notably, even the initial population of our FlowPIE surpasses strong baselines such as SCIPIP. 
We report the standard deviation (\textit{std}) for each task in Table~\ref{tab:main-results-ai-idea-bench} to provide a more reliable evaluation. The consistently lower \textit{std} of FlowPIE indicates that it generates ideas with greater robustness and more consistently high quality.
We further provide additional results in Appendix~\ref{appendix:effects-of-backbones} using \texttt{Qwen2.5-7B} and \texttt{LLaMA3.1-8B} as backbones, demonstrating that our method generalizes across different LLMs. Although the absolute performance is bounded by the capability of the underlying model, our method consistently yields stable relative improvements.

\paragraph{Reward Performance.} As shown in Table~\ref{tab:reward-results}, we evaluate all generated ideas on AI Idea Bench 2025 using the GRM, where the reward computation is aligned with the fitness evaluation in our evolution process. We use \texttt{DeepSeek-V3.2} model as the backbone of the GRM. The final idea population of FlowPIE achieves the highest average reward score among all baselines, with its initial idea population already outperforming competing methods. This initial population is generated using a threshold-based stopping criterion to guarantee sufficient exploration, instead of a fixed budget or step size.

Additionally, we visualize the reward lifecycle of FlowPIE in Figure~\ref{fig:reward-on-ai-idea-bench}, which exhibits a scaling trend in reward. The rewards of the initial ideas increase with evolution steps, although a bottleneck exists despite already being higher than other baseline methods. Subsequent idea evolution from this initial population substantially improves the rewards and shifts the distribution toward higher scores, as reflected by the continuously decreasing standard deviation. We conduct correlation analysis between reward and benchmark results in Appx.~\ref{appendix:reward-benchmark-consistency}.

\begin{wraptable}{r}{0.50\columnwidth}
\vspace{-1em}
 
\setlength{\tabcolsep}{0.7mm}
\scalebox{0.71}{
\begin{tabular}{lccccccccc}
\toprule
\textbf{Method}&\textbf{AI}&\textbf{HM}&\textbf{GMB} &\textbf{ES} &\textbf{NCS} &\textbf{TE} &\textbf{SBS} &\textbf{MS} &\textbf{PHP} \\
\midrule
\textbf{SCIPIP} &0.62&0.60&0.64&0.63&0.63&0.62&0.61&0.67&0.60 \\
\textbf{Research Agent} &0.58&0.63&0.66&0.65&0.65&0.63&0.60&0.69&0.58 \\
\textbf{Chain-of-Ideas} &0.63&0.67&\underline{0.71}&0.71&\underline{0.70}&\underline{0.70}&0.66&0.73&0.67 \\
\textbf{VirSci} &0.60&0.65&0.69&0.66&0.67&0.67&0.61&0.70&0.60 \\
\rowcolor{gray!25}\textbf{FlowPIE(\textit{Ours})} 
&\textbf{0.75}&\textbf{0.81}&\textbf{0.80}&\textbf{0.83}&\textbf{0.81}&\textbf{0.77}&\textbf{0.76}&\textbf{0.88}&\textbf{0.78} \\
\rowcolor{gray!15}\textbf{Initial Population}
&\underline{0.68}&\underline{0.72}&0.67&\underline{0.78}&\underline{0.70}&0.68&\underline{0.67}&\underline{0.83}&\underline{0.69} \\
\bottomrule
\end{tabular}}
\caption{Domain-specific reward performance across nine domains.}
\label{tab:domain}
\vspace{-1em}
\end{wraptable}
\noindent\textbf{Domain Generalization.} 
Due to the multi-domain nature of IdeaBench, which encompasses eight fields spanning Health \& Medicine~(HM), Genetics \& Molecular Biology~(GMB), Environmental Sciences~(ES), Neuroscience \& Cognitive Sciences~(NCS), Technology \& Engineering,~(TE) Social \& Behavioral Sciences~(SBS), Materials Science~(MS) and Public Health \& Policy~(PHP), together with the AI domain covered by AI Idea Bench 2025. We report domain-specific rewards in Table~\ref{tab:domain}, which demonstrate the strong domain generalization of FlowPIE. It achieves the highest rewards across nine different domains, especially excelling in Materials Science.

\begin{wraptable}{r}{0.5\columnwidth}
\vspace{-1em}
 
\setlength{\tabcolsep}{1.1mm}
\scalebox{0.85}{
\begin{tabular}{llcccc}
\toprule
\textbf{Categories} & \textbf{Metric} & \textbf{N} & \textbf{F} & \textbf{E} & \textbf{EE} \\
\midrule
Human and Human 
& Spearman $\rho$ 
& 0.49 & 0.54 & 0.32 & 0.47 \\
\midrule
Human and Reward 
& Spearman $\rho$  
& 0.60 & 0.87 & -- & -- \\
\bottomrule
\end{tabular}}
\caption{Inter-expert agreement of human evaluation and correlation between our reward and the average human score on novelty and feasibility.}
\label{tab:human-consistency}
\vspace{-1em}
\end{wraptable}
\paragraph{Human Evaluation Results.} We report the human evaluation results in Table~\ref{tab:reward-results}, demonstrating that the ideas generated by FlowPIE are of higher quality than those produced by the baselines.
Additionally, we report the inter-expert Spearman correlations for each metric, together with the correlation between average reward and average human score in Table~\ref{tab:human-consistency}.
Our reward demonstrates strong alignment with human judgments, achieving Spearman correlations of 0.60 and 0.87 with the average human ratings on novelty and feasibility, respectively. 
Notably, these correlations are comparable to or even higher than the agreement between individual experts, suggesting that the reward reliably captures aggregated human preferences.

\subsection{Analysis}

\begin{wrapfigure}{r}{0.47\columnwidth}
 
\includegraphics[width=1\linewidth]{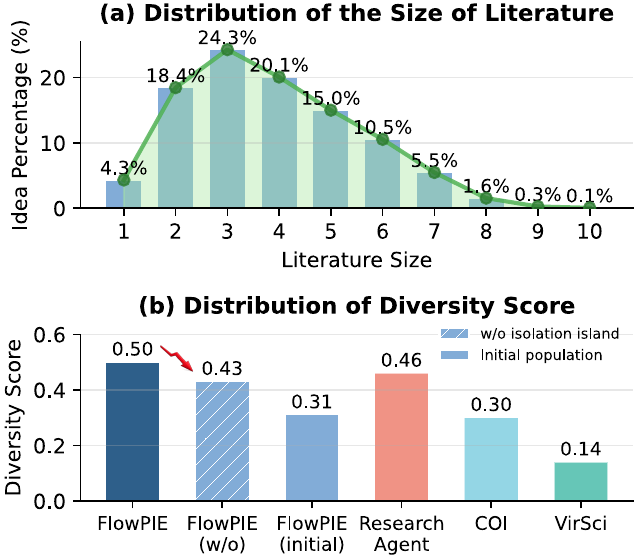}
\caption{(a) Distribution of explored literature count; (b) Diversity score distribution.}
\label{fig:size-literature}
\vspace{-3em}
\end{wrapfigure}

\noindent\textbf{Literature Exploration Analysis (RQ1).}
\label{literature-exploration}
Most prior works recognize the importance of retrieved literature for ideation. They adopt keyword–semantic hybrid retrieval strategies or construct entity-centric citation graphs to equip the framework with more relevant scientific knowledge. However, these methods primarily enrich the literature in a one-shot, static manner by retrieving a fixed number of papers. As shown in Figure~\ref{fig:size-literature}~(a), our FlowPIE couples retrieval with generation, enabling dynamically broader and deeper exploration. 
We observe that most ideas rely on approximately 3 publications in their trajectories, some incorporate more than seven or even ten during initial ideation. Notably, this variation is dynamically determined by ideas' quality rather than a hyperparameter.

\noindent\textbf{Idea Diversity Analysis (RQ2).}
\label{diversity-analysis}
We assess the diversity of generated ideas following~\citet{sican}, encoding them with all-MiniLM-L6-v2. We consider two ideas to be similar if their cosine similarity over threshold 0.65. The results are illustrated in Figure~\ref{fig:size-literature}~(b). 
\citet{sican} investigates idea diversity scaling with the increasing number of ideas at test-time, and find that simply increasing the quantity  leading to repeating duplicate ideas. Our FlowPIE achieves the superior diversity score and even the initial population's diversity outperforms Chain-of-Ideas and VirSci. This result demonstrates that dynamic literature exploration and isolated island–enhanced idea evolution can improve diversity beyond simple count scaling at test-time.

\subsection{Ablation Study}

\noindent\textbf{Ablations on flow-guided literature exploration.}
Prior LLM-enhanced EAs typically use zero-shot prompting or other methods' solutions as their initial population, but do not focus on optimizing the quality of their owns. 
For example, EvoAgent~\citep{yuan-etal-2025-evoagent} initializes its agent population using MetaGPT~\citep{hong2024metagpt}.
Our flow-guided MCTS dynamically explores the literature, filtering low-quality entries and weak ideas to build a high-quality population for subsequent evolution.
We conduct an ablation study by using ideas from SCIPIP and Chain-of-Ideas as the initial population, as reported in Table~\ref{tab:idea-evolution-on-baseline}, which demonstrates consistent performance improvements. 
This demonstrates the strong generalization of our idea evolution, as it effectively enhances ideas generated by heterogeneous methods while achieving the largest gains under our initialization strategy.

\begin{wraptable}{r}{0.55\columnwidth}
\vspace{-4mm}
 
\setlength{\tabcolsep}{0.8mm}
\scalebox{0.68}{
\begin{tabular}{lc|ccc}
\toprule
\textbf{Method} & \textbf{AR $\uparrow$} & \textbf{I2T $\uparrow$} & \textbf{I2I $\uparrow$} & \textbf{IMCQ $\uparrow$} \\
\midrule
\textbf{SCIPIP} & 0.61 & 4.18 & 3.68 & 0.464 \\
\text{ }{w/ Idea Evolution} & 0.63\imp{3.3\%} & 4.24~\imp{1.4\%} & 3.70~\imp{0.5\%} & 0.475~\imp{2.4\%} \\
\textbf{Chain-of-Ideas} & 0.59 & 4.21 & 3.62 & 0.437 \\
\text{ }{w/ Idea Evolution} & 0.64\imp{8.5\%} & 4.23~\imp{0.5\%} & 3.63~\imp{0.3\%} & 0.448~\imp{2.5\%} \\
\midrule
\textbf{FlowPIE (\textit{initial})} & 0.68 & 4.08 & 3.94 & 0.688 \\
\text{ }{w/ Idea Evolution} & 0.75~\imp{10.1\%} & 4.18~\imp{2.5\%} & 4.15~\imp{5.3\%} & 0.708~\imp{2.9\%} \\
\bottomrule
\end{tabular}}
\caption{Idea evolution results with initial ideas from different baselines. We report the average reward (AR) and average score of three tasks in AI Idea Bench 2025.}
\label{tab:idea-evolution-on-baseline}
\vspace{-6mm}
\end{wraptable}
\noindent\textbf{Ablations on idea evolution, \textit{i.e. just initialization}.} We conduct an ablation study in which removing the idea evolution process, and only using the flow-guided MCTS to generate the initial population. 
As shown in Table~\ref{tab:main-results-ai-idea-bench},~\ref{tab:ideabench}, and~\ref{tab:reward-results}, the initial idea population even outperforms other strong baselines, demonstrating the effectiveness of our method. However, the initial ideas exhibit a bottleneck despite increasing exploration iterations. Figure~\ref{fig:reward-on-ai-idea-bench} shows that increasing the number of exploration steps, corresponding to greater breadth and depth in graph, leads to marginal gains, with the rewards of the initial ideas approaching a plateau. 

\paragraph{Ablations on isolation islands within mutation.} We conduct an ablation study on the isolation islands within the mutation operator of idea evolution.
As shown in Figure~\ref{fig:size-literature}~(b), the diversity of ideas decreasing compared with full setting.
The reward results under this ablation setting (Table~\ref{tab:reward-results}, last row) only slightly underperform the full setting, suggesting that the isolation island strategy promotes diversity while preserving overall quality.

\begin{table*}[!t]
\small
\sethlcolor{orange!30} 
\scalebox{0.97}{\begin{tabular}{p{\linewidth}}
\toprule
\textbf{Input topic}: \textit{The topic of this paper is improving reasoning capabilities in Large Language Models (LLMs).} \\
\midrule
\textbf{Final Idea Title}: \textit{{Dynamic Macro-Guided Verification (DMGV): A Modular Reasoning Augmentation Framework for LLMs}}

\\\textbf{Summary}: The proposed Dynamic Macro-Guided Verification (DMGV), a framework that \hl{converts recurrent reasoning subchains in CoT traces into \textbf{reusable reasoning macros}, and attaches lightweight verifiers to validate their outputs}. Instead of repeatedly regenerating similar reasoning steps, the model composes solutions by invoking validated macros and verifying their outputs. The core insight is that many reasoning traces contain repeated sub-structures that can be compressed into reusable operators. By validating macro outputs through interface-level verification rather than re-evaluating entire reasoning chains, DMGV aims to \hl{reduce both error accumulation and token-level redundancy in LLM reasoning}. It consists of 3 stages: 
\textit{(1) Macro discovery}: reasoning traces are segmented and clustered to identify recurrent subchains, which are distilled into parameterized macros with explicit input-output interfaces.
\textit{(2) Macro-guided reasoning}: during inference, the generator composes reasoning steps by invoking macros whenever a subgoal matches a macro signature.
\textit{(3) Targeted verification}: a lightweight verifier checks macro outputs using symbolic, arithmetic, or structural consistency tests; failed cases trigger localized fallback re-synthesis.
\\
\bottomrule
\end{tabular}}
\caption{Demonstration of an idea generated by FlowPIE. Due to space limitations, we present only a concise summary of the final idea. The complete version of this idea is provided in Table~\ref{case:full-demo}.}
\label{example:case-study}
\end{table*}

\subsection{Qualitative Analysis}
We present an interesting case study on reasoning pattern abstraction for LLM reasoning in Table~\ref{example:case-study}, showing how recurrent CoT subchains can be transformed into reusable and verifiable reasoning units, termed as \textit{reasoning macros}, which aims to reduce both error accumulation and token-level redundancy in LLM reasoning. Unlike prior modular reasoning approaches, this idea equips each macro with an explicit interface and lightweight verification, enabling reliable reuse, reducing repeated low-level generation, and localizing failures to individual macro calls. Additional cases across various domains are in Appendix~\ref{appendix:case}.

\section{Conclusion}
In this paper, we rethink the traditional scientific idea generation that statically retrieves literature before generating ideas, and propose a novel test-time idea evolution framework, FlowPIE. It integrates initial idea generation with dynamic literature exploration via flow-guided MCTS, along with a variety of evolutionary operators for continuous idea refinement. 
We combine literature retrieval with feedback on the quality of initial ideas, assessed via GRM-based rewards, to enable adaptive exploration by adjusting the flow probability. 
Human evaluation and benchmark results demonstrate that FlowPIE generates ideas that are more relevant to the query, higher-quality, and more stable than those produced by other strong baselines, while also exhibiting strong domain generalization across various scientific fields. 
Beyond benchmarks, we conduct a reward analysis, which demonstrates clear test-time scaling of rewards in test time. 

\bibliography{ref}

@inproceedings{chen2024m3,
  title={M3-embedding: Multi-linguality, multi-functionality, multi-granularity text embeddings through self-knowledge distillation},
  author={Chen, Jianlyu and Xiao, Shitao and Zhang, Peitian and Luo, Kun and Lian, Defu and Liu, Zheng},
  booktitle={Findings of the association for computational linguistics: ACL 2024},
  pages={2318--2335},
  year={2024},
  url={https://aclanthology.org/2024.findings-acl.137.pdf}
}

@article{liu2024deepseek,
  title={Deepseek-v3 technical report},
  author={Liu, Aixin and Feng, Bei and Xue, Bing and Wang, Bingxuan and Wu, Bochao and Lu, Chengda and Zhao, Chenggang and Deng, Chengqi and Zhang, Chenyu and Ruan, Chong and others},
  journal={arXiv preprint arXiv:2412.19437},
  year={2024},
  url={https://arxiv.org/abs/2412.19437}
}

@article{wang2024scipip,
  title={Scipip: An llm-based scientific paper idea proposer},
  author={Wang, Wenxiao and Gu, Lihui and Zhang, Liye and Luo, Yunxiang and Dai, Yi and Shen, Chen and Xie, Liang and Lin, Binbin and He, Xiaofei and Ye, Jieping},
  journal={arXiv preprint arXiv:2410.23166},
  year={2024},
  url={https://arxiv.org/abs/2410.23166}
}

@inproceedings{
guo2024connecting,
title={Connecting Large Language Models with Evolutionary Algorithms Yields Powerful Prompt Optimizers},
author={Qingyan Guo and Rui Wang and Junliang Guo and Bei Li and Kaitao Song and Xu Tan and Guoqing Liu and Jiang Bian and Yujiu Yang},
booktitle={The Twelfth International Conference on Learning Representations},
year={2024},
url={https://openreview.net/forum?id=ZG3RaNIsO8}
}

@inproceedings{yuan-etal-2025-evoagent,
    title = "{E}vo{A}gent: Towards Automatic Multi-Agent Generation via Evolutionary Algorithms",
    author = "Yuan, Siyu  and
      Song, Kaitao  and
      Chen, Jiangjie  and
      Tan, Xu  and
      Li, Dongsheng  and
      Yang, Deqing",
    editor = "Chiruzzo, Luis  and
      Ritter, Alan  and
      Wang, Lu",
    booktitle = "Proceedings of the 2025 Conference of the Nations of the Americas Chapter of the Association for Computational Linguistics: Human Language Technologies (Volume 1: Long Papers)",
    month = apr,
    year = "2025",
    address = "Albuquerque, New Mexico",
    publisher = "Association for Computational Linguistics",
    url = "https://aclanthology.org/2025.naacl-long.315/",
    doi = "10.18653/v1/2025.naacl-long.315",
    pages = "6192--6217",
    ISBN = "979-8-89176-189-6",
}

@ARTICLE{2025arXiv250109891L,
       author = {{Lee}, Kuang-Huei and {Fischer}, Ian and {Wu}, Yueh-Hua and {Marwood}, Dave and {Baluja}, Shumeet and {Schuurmans}, Dale and {Chen}, Xinyun},
        title = "{Evolving Deeper LLM Thinking}",
      journal = {arXiv e-prints},
         year = 2025,
        month = jan,
          eid = {arXiv:2501.09891},
        pages = {arXiv:2501.09891},
          doi = {10.48550/arXiv.2501.09891},
archivePrefix = {arXiv},
       eprint = {2501.09891},
 primaryClass = {cs.AI},
       adsurl = {https://ui.adsabs.harvard.edu/abs/2025arXiv250109891L}
}

@inproceedings{li2024chain,
    title = "Chain of Ideas: Revolutionizing Research Via Novel Idea Development with {LLM} Agents",
    author = "Li, Long  and
      Xu, Weiwen  and
      Guo, Jiayan  and
      et al.",
    editor = "Christodoulopoulos, Christos  and
      Chakraborty, Tanmoy  and
      Rose, Carolyn  and
      Peng, Violet",
    booktitle = "Findings of the Association for Computational Linguistics: EMNLP 2025",
    month = nov,
    year = "2025",
    address = "Suzhou, China",
    publisher = "Association for Computational Linguistics",
    url = "https://aclanthology.org/2025.findings-emnlp.477/",
    doi = "10.18653/v1/2025.findings-emnlp.477",
    pages = "8971--9004",
    ISBN = "979-8-89176-335-7",
}

@article{hurst2024gpt,
  title={Gpt-4o system card},
  author={Hurst, Aaron and Lerer, Adam and Goucher, Adam P and Perelman, Adam and Ramesh, Aditya and Clark, Aidan and Ostrow, AJ and Welihinda, Akila and Hayes, Alan and Radford, Alec and others},
  journal={arXiv preprint arXiv:2410.21276},
  year={2024},
  url={https://arxiv.org/abs/2410.21276}
}

@inproceedings{baek2025researchagent,
  title={Researchagent: Iterative research idea generation over scientific literature with large language models},
  author={Baek, Jinheon and Jauhar, Sujay Kumar and Cucerzan, Silviu and Hwang, Sung Ju},
  booktitle={Proceedings of the 2025 Conference of the Nations of the Americas Chapter of the Association for Computational Linguistics: Human Language Technologies (Volume 1: Long Papers)},
  pages={6709--6738},
  year={2025},
  url={https://aclanthology.org/2025.naacl-long.342/}
}

@inproceedings{su2025many,
  title={Many heads are better than one: Improved scientific idea generation by a llm-based multi-agent system},
  author={Su, Haoyang and Chen, Renqi and Tang, Shixiang and Yin, Zhenfei and Zheng, Xinzhe and Li, Jinzhe and Qi, Biqing and Wu, Qi and Li, Hui and Ouyang, Wanli and others},
  booktitle={Proceedings of the 63rd Annual Meeting of the Association for Computational Linguistics (Volume 1: Long Papers)},
  pages={28201--28240},
  year={2025},
  url={https://aclanthology.org/2025.acl-long.1368/}
}

@article{RomeraParedes2023MathematicalDF,
  title={Mathematical discoveries from program search with large language models},
  author={Bernardino Romera-Paredes and Mohammadamin Barekatain and Alexander Novikov and Matej Balog and M. Pawan Kumar and Emilien Dupont and Francisco J. R. Ruiz and Jordan S. Ellenberg and Pengming Wang and Omar Fawzi and Pushmeet Kohli and Alhussein Fawzi and Josh Grochow and Andrea Lodi and Jean-Baptiste Mouret and Talia Ringer and Tao Yu},
  journal={Nature},
  year={2023},
  volume={625},
  pages={468 - 475},
  url={https://api.semanticscholar.org/CorpusID:266223700}
}

@inproceedings{ye2025physics,
title={Physics of Language Models: Part 2.1, Grade-School Math and the Hidden Reasoning Process},
author={Tian Ye and Zicheng Xu and Yuanzhi Li and Zeyuan Allen-Zhu},
booktitle={The Thirteenth International Conference on Learning Representations},
year={2025},
url={https://openreview.net/forum?id=Tn5B6Udq3E}
}

@inproceedings{liao-etal-2024-medcare,
    title = "{M}ed{C}are: Advancing Medical {LLM}s through Decoupling Clinical Alignment and Knowledge Aggregation",
    author = "Liao, Yusheng  and
      Jiang, Shuyang  and
      Chen, Zhe  and
      Wang, Yu  and
      Wang, Yanfeng",
    editor = "Al-Onaizan, Yaser  and
      Bansal, Mohit  and
      Chen, Yun-Nung",
    booktitle = "Findings of the Association for Computational Linguistics: EMNLP 2024",
    month = nov,
    year = "2024",
    address = "Miami, Florida, USA",
    publisher = "Association for Computational Linguistics",
    url = "https://aclanthology.org/2024.findings-emnlp.619/",
    doi = "10.18653/v1/2024.findings-emnlp.619",
    pages = "10562--10581"
}

@ARTICLE{2025arXiv250701903C,
       author = {{Chen}, Qiguang and {Yang}, Mingda and {Qin}, Libo and {Liu}, Jinhao and {Yan}, Zheng and {Guan}, Jiannan and {Peng}, Dengyun and {Ji}, Yiyan and {Li}, Hanjing and {Hu}, Mengkang and {Zhang}, Yimeng and {Liang}, Yihao and {Zhou}, Yuhang and {Wang}, Jiaqi and {Chen}, Zhi and {Che}, Wanxiang},
        title = "{AI4Research: A Survey of Artificial Intelligence for Scientific Research}",
      journal = {arXiv e-prints},
         year = 2025,
        month = jul,
          eid = {arXiv:2507.01903},
        pages = {arXiv:2507.01903},
          doi = {10.48550/arXiv.2507.01903},
archivePrefix = {arXiv},
       eprint = {2507.01903},
 primaryClass = {cs.CL},
       adsurl = {https://ui.adsabs.harvard.edu/abs/2025arXiv250701903C}
}

@inproceedings{
tang2025airesearcherautonomousscientificinnovation,
title={{AI}-Researcher: Autonomous Scientific Innovation},
author={Jiabin Tang and Lianghao Xia and Zhonghang Li and Chao Huang},
booktitle={The Thirty-ninth Annual Conference on Neural Information Processing Systems},
year={2025},
url={https://openreview.net/forum?id=kQWyOYUAC4}
}

@inproceedings{sican,
title={Can {LLM}s Generate Novel Research Ideas? A Large-Scale Human Study with 100+ {NLP} Researchers},
author={Chenglei Si and Diyi Yang and Tatsunori Hashimoto},
booktitle={The Thirteenth International Conference on Learning Representations},
year={2025},
url={https://openreview.net/forum?id=M23dTGWCZy}
}

@misc{qiu2025aiideabench2025,
      title={AI Idea Bench 2025: AI Research Idea Generation Benchmark}, 
      author={Yansheng Qiu and Haoquan Zhang and Zhaopan Xu and Ming Li and Diping Song and Zheng Wang and Kaipeng Zhang},
      year={2025},
      eprint={2504.14191},
      archivePrefix={arXiv},
      primaryClass={cs.AI},
      url={https://arxiv.org/abs/2504.14191}, 
}

@article{lu2024ai,
	author = {Lu, Chris and Lu, Cong and Lange, Robert Tjarko and Yamada, Yutaro and Hu, Shengran and Foerster, Jakob and Ha, David and Clune, Jeff},
	date = {2026/03/01},
	doi = {10.1038/s41586-026-10265-5},
	id = {Lu2026},
	isbn = {1476-4687},
	journal = {Nature},
	number = {8107},
	pages = {914--919},
	title = {Towards end-to-end automation of AI research},
	url = {https://doi.org/10.1038/s41586-026-10265-5},
	volume = {651},
	year = {2026}}

@inproceedings{guo2025ideabench,
  title={Ideabench: Benchmarking large language models for research idea generation},
  author={Guo, Sikun and Shariatmadari, Amir Hassan and Xiong, Guangzhi and Huang, Albert and Kim, Myles and Williams, Corey M and Bekiranov, Stefan and Zhang, Aidong},
  booktitle={Proceedings of the 31st ACM SIGKDD Conference on Knowledge Discovery and Data Mining V. 2},
  pages={5888--5899},
  year={2025},
  url={https://dl.acm.org/doi/10.1145/3711896.3737419}
}

@article{xu2025probing,
  title={Probing Scientific General Intelligence of LLMs with Scientist-Aligned Workflows},
  author={Xu, Wanghan and Zhou, Yuhao and Zhou, Yifan and Cao, Qinglong and Li, Shuo and Bu, Jia and Liu, Bo and Chen, Yixin and He, Xuming and Zhao, Xiangyu and others},
  journal={arXiv preprint arXiv:2512.16969},
  year={2025},
  url={https://arxiv.org/abs/2512.16969}
}

@inproceedings{
hong2024metagpt,
title={Meta{GPT}: Meta Programming for A Multi-Agent Collaborative Framework},
author={Sirui Hong and Mingchen Zhuge and Jonathan Chen and Xiawu Zheng and Yuheng Cheng and Jinlin Wang and Ceyao Zhang and Zili Wang and Steven Ka Shing Yau and Zijuan Lin and Liyang Zhou and Chenyu Ran and Lingfeng Xiao and Chenglin Wu and J{\"u}rgen Schmidhuber},
booktitle={The Twelfth International Conference on Learning Representations},
year={2024},
url={https://openreview.net/forum?id=VtmBAGCN7o}
}

@article{Bengio2021GFlowNetF,
  title={Gflownet foundations},
  author={Bengio, Yoshua and Lahlou, Salem and Deleu, Tristan and Hu, Edward J and Tiwari, Mo and Bengio, Emmanuel},
  journal={Journal of Machine Learning Research},
  volume={24},
  number={210},
  pages={1--55},
  year={2023},
  url={https://jmlr.org/papers/v24/22-0364.html}
}

\appendix

\section{More Related Work Discussion}
\label{sec:more-related-work}

\paragraph{Scientific Idea Generation Evaluation and Benchmarks.}
As LLMs continue to demonstrate escalating performance across diverse tasks, there has been a notable proliferation of specialized frameworks designed for idea generation, To effectively measure the efficacy of these systems, several rigorous benchmarks, such as Scientist-Bench~\citep{tang2025airesearcherautonomousscientificinnovation}, AI Idea Bench 2025~\citep{qiu2025aiideabench2025}, IdeaBench~\citep{guo2025ideabench} and SGI-Bench~\cite{xu2025probing}. 
These benchmarks provide the necessary empirical groundwork to validate the substantive quality of proposed ideas, ensuring they are assessed not only for their surface-level novelty but also for their underlying feasibility and technological grounding.

In this paper, we adopt AI Idea Bench 2025 and Ideabench due to their suitability for our specific idea generation task and their inclusion of multi-domain attributes. These two benchmarks derive their ideas from recent high-quality publications, such as top conferences in the AI domain, providing golden reference ideas. In contrast, SGI-Bench is based on \textit{Science’s 125 Big Questions}, which do not have explicit optimal idea answers. Additionally, SGI-Bench provides the literature related to each idea in advance, which limits the evaluation of our method’s literature exploration ability. For Scientist-Bench, idea generation is only a subset or starting point of its overall task, which focuses on a full-stage automated research pipeline, including experiment execution, code generation, and paper writing, differing from our research objective.

Due to the complexity and subjective nature of evaluating idea generation, these benchmarks mainly include two types of metrics: LLM-as-a-judge paradigm scores and similarity-based scores. For our experiments on AI Idea Bench 2025 and IdeaBench, which provide golden idea references, we can compute similarity-based metrics, such as I2T and I2I tasks in AI Idea Bench 2025, and Semantic Similarity and Idea Overlap in IdeaBench. They also provide LLM-as-a-judge metrics, such as the IMCQ task in AI Idea Bench 2025 and the Novelty Insight Score and Feasibility Insight Score in IdeaBench. These metrics do not rely on a single scalar score from an LLM; instead, they evaluate the relative quality of generated ideas from pairwise and listwise perspectives. 

Although SGI-Bench proposes a combined objective and subjective metric for this task, its objective component heavily relies on the provided related literature. This makes it unfair for frameworks that can retrieve literature themselves, such as SCIPIP, Chain-of-Thought, and our FlowPIE. The objective metric is more suitable for evaluating the scientific information integration ability of the backbone model.

Additionally, the idea formats in prior works vary significantly. For example, Chain-of-Ideas~\citep{li2024chain} structures ideas with a title, motivation, method, and experimental plan, whereas SCIPIP~\citep{wang2024scipip} includes only problem and method sections. Our work builds on the Chain-of-Ideas and Research Agent~\citep{baek2025researchagent} formats, including motivation, method, and experimental plan. These different idea formats also cause the evaluation to be unstable. We believe that unifying the idea format and constructing standard evaluation metrics for these idea generation benchmarks is very important. Even though our evaluation setup includes GRM-based rewards, we consider this standardization an important direction for future work in the community.

\paragraph{Evolutionary Algorithm with LLMs.} Evolutionary algorithms (EAs) are usually used for optimization problem search. With the advancement of LLMs, they can be leveraged as both an optimizer and a generator in EAs, thanks to their extensive out-of-domain knowledge and ability to operate in semantically rich language spaces. Researchers have already explored incorporating LLMs into EAs frameworks. 
\citet{guo2024connecting} propose EvoPrompt, which combines an LLM-enhanced evolutionary algorithm framework with discrete prompt optimization.
\citet{yuan-etal-2025-evoagent} propose the EvoAgent framework, which treats existing agent frameworks as initial individuals and applies a series of evolutionary operators, significantly improving performance on writing, code, and travel planning tasks.
Notably, \citet{2025arXiv250109891L} introduce a framework called Mind Evolution, which leverages evolutionary algorithms to scale LLMs’ test-time computation, outperforming Best-of-N and Sequential Revision test-time reasoning strategies at the same computational cost.

Our FlowPIE builds on EAs for test-time computation, leveraging multiple operators to refine the idea population based on the LLM-based idea generator. Additionally, prior works do not focus on evolving a high-quality initial population. Most use zero-shot results as the starting population or rely on results from other methods, such as EvoAgent. We argue that successful evolution depends on a strong initial parent idea population. Therefore, we propose a novel method for constructing the initial idea population, combining dynamic literature exploration and initial idea generation with GFlowNet-inspired flow-guided MCTS, balancing exploration and exploitation of the literature.

\section{Experimental Setting}

\subsection{Implementation Details}
\label{appendix:implementation-details}

\begin{wraptable}{r}{0.5\columnwidth}
\vspace{-1em}
 
\setlength{\tabcolsep}{1mm}
\scalebox{0.67}{
\begin{tabular}{lccc}
\toprule
\textbf{Method} & \textbf{Prompt Tokens} & \textbf{Avg. Tokens / Idea} & \textbf{Avg. Cost / Idea} \\
\midrule
\textbf{SCIPIP} & 4062.50 & 921.33 & \$0.038 \\
\textbf{Research Agent} & 1755.44 & 477.53 & \$0.028 \\
\textbf{Chain-of-Ideas} & 4526.92 & 650.43 & \$0.052 \\
\textbf{Virsci} & 1458.36 & 789.60 & \$0.061 \\
\textbf{FlowPIE (\textit{Ours})} & 4123.43 & 924.25 & \$0.056 \\
\bottomrule
\end{tabular}}
\caption{API cost comparison for idea generation across different methods. All methods use GPT-4o-mini as the idea generator.}
\label{tab:api_cost}
\vspace{-1em}
\end{wraptable}
 We implement our FlowPIE using publicly available LLMs for both idea generation and evaluation. Specially, we use \texttt{GPT-4o-mini} as the backbone model for idea generator, accessed through the official API. For fitness evaluation and idea quality analysis, we employ \texttt{DeepSeek-V3.2} as the fitness evaluator.
For flow-guided MCTS, the exploration coefficient \(c\) in Eq.~\ref{algorithm-1} is set to commonly used value \(c=\sqrt{2}\) for balancing exploration and exploitation. The learning rate $\alpha$ for updating the flow probability $P_{flow}$ in Eq.~\ref{algorithm-2} is set to 0.2. For the test-time Idea Evolution, the maximum number of evolution iterations is limited to 20. Additionally, if the standard deviation of the generated ideas' fitness score is lower than threshold 0.05, the evolution process terminates as the final ideas. 

\begin{algorithm*}[!t]
\small
\caption{Initial Idea Population Construction via Flow-Guided MCTS}
\label{algorithm-1}
\begin{algorithmic}[1]
\KwIn{Query $q$, patent literature graph $\mathcal{G}=(\mathcal{V},\mathcal{E})$, maximum iteration $N$, convergence threshold $\epsilon$}
\KwOut{Initial idea set $\mathcal{I}$}

\State Initialize root node $s_0 \leftarrow q$, global flow $F(s_0)\leftarrow 1$ 
~{\textcolor{commentcolor}{~~~~\text{// Treat the query as the root of literature exploration.}}}

\State Retrieve top-$k$ relevant patents as children of $s_0$ 
~{\textcolor{commentcolor}{~~~~\text{// Initialize the starting literature nodes via semantic retrieval.}}}

\State Initialize $\mathcal{I}\leftarrow\varnothing$, reward buffer $\mathcal{R}\leftarrow\varnothing$, $n\leftarrow 0$

\While{$n < N$ \textbf{and} $\mathrm{Var}(\mathcal{R}) > \epsilon$}
    \State $n \leftarrow n+1$, $s \leftarrow s_0$, $\phi \leftarrow [s_0]$ 
    ~{\textcolor{commentcolor}{~~~~\text{// Start a new literature exploration trajectory.}}}
    
    \While{$A(s)\neq \varnothing$}
        \State $s' \leftarrow \arg\max\limits_{u\in A(s)} UCB(u\mid s)$ 
        ~{\textcolor{commentcolor}{~~~~\text{// Select next node via flow-guided UCB (Eq.~\ref{UCB}).}}}
        \State Append $s'$ to $\phi$, set $s \leftarrow s'$
        
        \If{$s$ is expandable and not fully explored}
            \State Expand one adjacent patent node into $A(s)$
            ~{\textcolor{commentcolor}{~~~~\text{// Expand a new literature node for exploration.}}}
            \State \textbf{break}
        \EndIf
    \EndWhile
    
    \State $i \leftarrow \text{GenerateIdea}(\phi)$ 
    ~{\textcolor{commentcolor}{~~~~\text{// Generate a scientific idea conditioned on the literature trajectory.}}}

    \State $R \leftarrow \text{GRM}(i)$ 
    ~{\textcolor{commentcolor}{~~~~\text{// Evaluate the idea using the generative reward model.}}}

    \State $\mathcal{I} \leftarrow \mathcal{I} \cup \{i\}$, $\mathcal{R} \leftarrow \mathcal{R} \cup \{R\}$

    \For{each transition $(s_t \rightarrow s_{t+1})$ in $\phi$}
        \State Compute $\tilde{R}_t = R\cdot \gamma^{T-t}$ 
        ~{\textcolor{commentcolor}{~~~~\text{// Apply depth-aware reward decay.}}}

        \State Update $Q(s_{t+1}\mid s_t)$ by running average
        \State Update $P_f(s_{t+1}\mid s_t)$ using Eq.~\ref{flow}

        \State Normalize $P_f(\cdot\mid s_t)$ over $A(s_t)$
        \State Update global flow $F(s_{t+1}) = F(s_t)\cdot P_f(s_{t+1}\mid s_t)$ 
        ~{\textcolor{commentcolor}{~~~~\text{// Maintain consistency with global flow constraint.}}}
    \EndFor

\EndWhile

\State \Return $\mathcal{I}$ 
~{\textcolor{commentcolor}{~~~~\text{// Generated ideas form the initial population for the evolution stage.}}}

\end{algorithmic}
\end{algorithm*}

\paragraph{Method Cost.}
We report the API cost of each method for generating a scientific idea, where \texttt{GPT-4o-mini} is used as the LLM-based idea generator across all methods, as shown in Table~\ref{tab:api_cost}.

\subsection{Patent-based Literature Database Details}
\label{appendix:patent-literature}
\begin{wraptable}{r}{0.4\columnwidth}
\vspace{-1em}
\scalebox{0.8}{
\begin{tabular}{lcc}
\toprule
\textbf{Section} & \textbf{Count} & \textbf{Percentage (\%)} \\
\midrule
A & 1293 & 8.06 \\
B & 980 & 6.11 \\
C & 710 & 4.42 \\
D & 26 & 0.16 \\
E & 123 & 0.77 \\
F & 208 & 1.30 \\
G & 7957 & 49.58 \\
H & 4753 & 29.61 \\
\midrule
\textbf{All} & \textbf{16050} & \textbf{100} \\
\bottomrule
\end{tabular}}
\caption{Distribution of IPC sections.}
\label{tab:all_ipc}
\vspace{-1em}
\end{wraptable}
\paragraph{Literature Source and Data Statistics.} 
Our patent-based literature dataset is constructed from available resources released by the USPTO\footnote{https://data.uspto.gov/home}. We collect 16,050 patents published between October and December 2025. The selected patents cover a wide range of domains organized under the International Patent Classification (IPC) system, ensuring diversity across different technical fields and innovation areas. We provide the distribution of patents across major IPC sections in our dataset in Table~\ref{tab:all_ipc}, which illustrates the technological coverage of the constructed patent corpus.

\begin{wrapfigure}{r}{0.52\columnwidth}
 
\includegraphics[width=\linewidth]{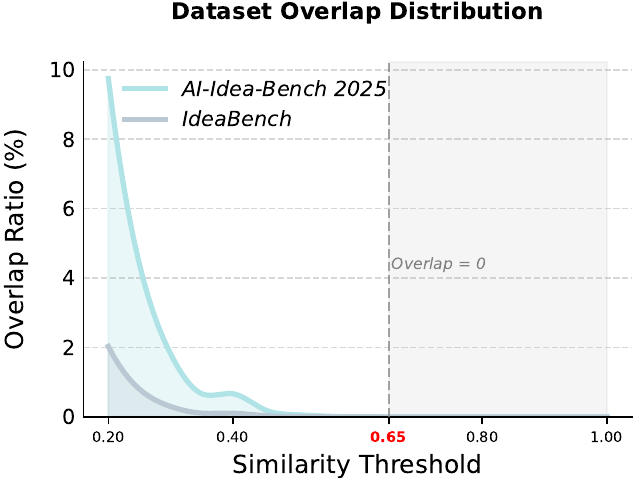}
\caption{Dataset overlap distribution between our literature and two benchmarks across different similarity levels. The overlap ratio rapidly decreases as similarity increases and approaches zero in the high-similarity region, indicating minimal risk of data leakage.}
\label{fig:data-leakage}
\vspace{-2em}
\end{wrapfigure}
\paragraph{Data Collection and Processing.} 
The raw patent documents are first parsed into structured textual components, including abstracts, claims, and citation metadata. To facilitate structured reasoning over the patent corpus, we employ a LLM to extract key semantic attributes from each patent.

Following the formulation introduced in Section~\ref{initial-ideation}, each patent $p \in \mathcal{D}$ is mapped into an attribution tuple $\langle \mathcal{A}_p, \mathcal{F}_p, \mathcal{S}_p \rangle$. Where $\mathcal{A}_p$ represents its abstract, is directly extracted from the structured patent metadata. The $\mathcal{F}_p$ represents the core technical feature, which is generated by a LLM through analyzing the full patent. The extracted features are further verified by human experts to ensure their correctness and completeness. The $\mathcal{S}_p$ denotes its semantic embedding, which is obtained by encoding the patent using  \texttt{BGE-M3}~\citep{chen2024m3} as the embedding model.

\paragraph{Data Quality Analysis.}
We further perform data quality analyses to confirm that our patent-based literature database does not suffer from knowledge leakage.
Following \citet{sican}, we encode all literature (papers and patents) using \texttt{all-MiniLM-L6-v2} and compute the cosine similarity between our patent-based literature corpus and the target papers in the two benchmarks.
We provide the similarity distribution as shown in Figure~\ref{fig:data-leakage}. 
Most similarity scores fall below the threshold 0.3, with none patent approaching the predefined duplication threshold~(0.65). This indicates limited semantic overlap between the two corpora, suggesting that our literature database is largely independent of the benchmark datasets and poses minimal risk of data leakage.

\section{Methodology Details}

In this section, we provide further details on the motivation and formulation of the proposed flow-guided MCTS, including the flow-guided UCB for forward selection and the backpropagation mechanism.
The detailed algorithms are summarized in Algorithm~\ref{algorithm-1} and Algorithm~\ref{algorithm-2}.

\subsection{Flow-Guided Exploration Formulation}
We formulate literature exploration as a sequential retrieval process represented by a search tree. Each node corresponds to a partial state $s$, and each edge represents an expansion action producing a successor state $s' \in A(s)$.

\begin{algorithm*}[!t]
\small
\caption{Test-Time Idea Evolution}
\label{algorithm-2}
\begin{algorithmic}[1]
\KwIn{Initial idea population $\mathcal{I}_0$ of size $N$, literature graph $\mathcal{G}$, offspring number $M$, mutation rate $\rho$, maximum evolution step $T$, convergence threshold $\epsilon$}
\KwOut{Final evolved idea population $\mathcal{I}^{*}$}

\State Initialize population $\mathcal{I} \leftarrow \mathcal{I}_0$, evolution step $t \leftarrow 0$
~{\textcolor{commentcolor}{~~~~\text{// Use the ideas from Alg. 1 as the starting population.}}}

\State Evaluate all ideas in $\mathcal{I}$ with GRM to obtain fitness scores
~{\textcolor{commentcolor}{~~~~\text{// Fitness computation on novelty and feasibility.}}}

\While{$t < T$ \textbf{and} $\Delta\mathrm{Reward}(\mathcal{I}) > \epsilon$}
    \State $t \leftarrow t + 1$, $\mathcal{O} \leftarrow \varnothing$
    ~{\textcolor{commentcolor}{~~~~\text{// Generate a new offspring set at each evolution step.}}}
    
    \For{$m = 1$ to $M$}
        \State Select two parent ideas $i_a, i_b$ from $\mathcal{I}$
        \State $o \leftarrow \text{Crossover}(i_a, i_b)$
        ~{\textcolor{commentcolor}{~~~~\text{// Recombine core technical features of two promising parent ideas.}}}
        
        \If{$\text{Bernoulli}(\rho)$}
            \State $\mathcal{L}_{ood} \leftarrow \text{IsolationIsland}(\mathcal{G}, i_a, i_b)$
            \State $o \leftarrow \text{Mutate}(o, \mathcal{L}_{ood})$
            ~{\textcolor{commentcolor}{~~~~\text{// Inject literature from isolated island to encourage conceptual diversity.}}}
        \EndIf
        
        \State Evaluate offspring $o$ with GRM to obtain fitness $R(o)$
        \State $\mathcal{O} \leftarrow \mathcal{O} \cup \{o\}$
    \EndFor
    
    \State $\mathcal{C} \leftarrow \mathcal{I} \cup \mathcal{O}$
    ~{\textcolor{commentcolor}{~~~~\text{// Merge parent and offspring ideas into a candidate pool.}}}
    
    \State $\mathcal{I}_{next} \leftarrow \varnothing$
    \While{$|\mathcal{I}_{next}| < N$}
        \State Sample a random subset $\mathcal{S} \subset \mathcal{C}$
        \State $i^{*} \leftarrow \arg\max_{i \in \mathcal{S}} R(i)$
        \State $\mathcal{I}_{next} \leftarrow \mathcal{I}_{next} \cup \{i^{*}\}$
        \State $\mathcal{C} \leftarrow \mathcal{C} \setminus \{i^{*}\}$
        ~{\textcolor{commentcolor}{~~~~\text{// Tournament selection preserves high-fitness ideas for the next generation.}}}
    \EndWhile
    
    \State $\mathcal{I} \leftarrow \mathcal{I}_{next}$
\EndWhile

\State \Return $\mathcal{I}$
~{\textcolor{commentcolor}{~~~~\text{// Return the final evolved population after convergence.}}}
\end{algorithmic}
\end{algorithm*}

In standard MCTS, action selection relies on empirical value estimation: 
\begin{equation}
    Q(s'|s)=\frac{1}{N(s'|s)}\sum_{i=1}^{N(s'|s)}R_i
\end{equation}
where $N(s)$ and $N(s'|s)$ denote node and edge visitation counts, respectively. Classical UCB-based selection encourages exploration via uncertainty-driven bonuses:
\begin{equation}
    UCB(s'|s)=Q(s'|s)+c\sqrt{\frac{lnN(s)}{1+N(s'|s)}}
\end{equation}
while effective for optimization problems, this mechanism typically concentrates probability mass onto a single dominant trajectory.
However, scientific idea generation requires discovering multiple high-quality solutions rather than a single optimum.

To address this limitation, our FlowPIE introduces a global flow variable that redistributes exploration probability across the search tree.
FlowPIE associate each state with a non-negative flow quantity $F(s)$, representing the amount of exploration mass reaching the state. The root node is initialized as $F(s_0) =1$ The outgoing flow is decomposed over candidate expansions:
\begin{equation}
    F(s'|s)=F(s)P_f(s'|s)
\end{equation}
where $P_f(s'|s)$ denotes the relative importance of transition $s\to s'$. This induces the conservation constraint:
\begin{equation}
    \sum_{s_i \in A(s)}P_f(s_i|s) =1
\end{equation}

Therefore, we define a probability distribution governing exploration allocation. Replacing uniform exploration in classical UCB with flow-aware allocation yields the proposed \textbf{Flow-Guided UCB}:
\begin{equation}
\small
    UCB(s'|s)=Q(s'|s) + c\cdot P_f(s'|s)\frac{\sqrt{N(s)}}{1+N(s'|s)}
\end{equation}
Intuitively, transitions receiving larger global flow probability obtain stronger exploration preference, allowing promising reasoning directions to be revisited without collapsing exploration diversity.
The flow probability propagates recursively along trajectories $F(s_{i+1})=F(s_i)P_f(s_{i+1}|s_i)$, enabling downstream states to inherit global importance.

\subsection{Reward-Driven Flow Update}
Given a rollout terminating with reward $R$, temporal credit assignment is performed using a decay mechanism $\tilde{R}_t = R  \cdot \gamma ^{T-t}$, where $T$ denotes the maximum depth of trajectory, $\gamma$ is a discount factor. Transition flow is updated using exponential averaging:
\begin{equation}
     P_{f}(s'|s)\leftarrow(1-\alpha)P_{f}(s'|s) + \alpha \tilde{R}_t
\end{equation}
Under this update rule, transitions belonging to high-reward trajectories continuously accumulate larger incoming flow mass.

Consequently, the proposed search process approximates the core principle of Generative Flow Networks, encouraging sampling from a distribution of high-reward solutions rather than converging to a single optimal path.

\begin{figure*}[!h]
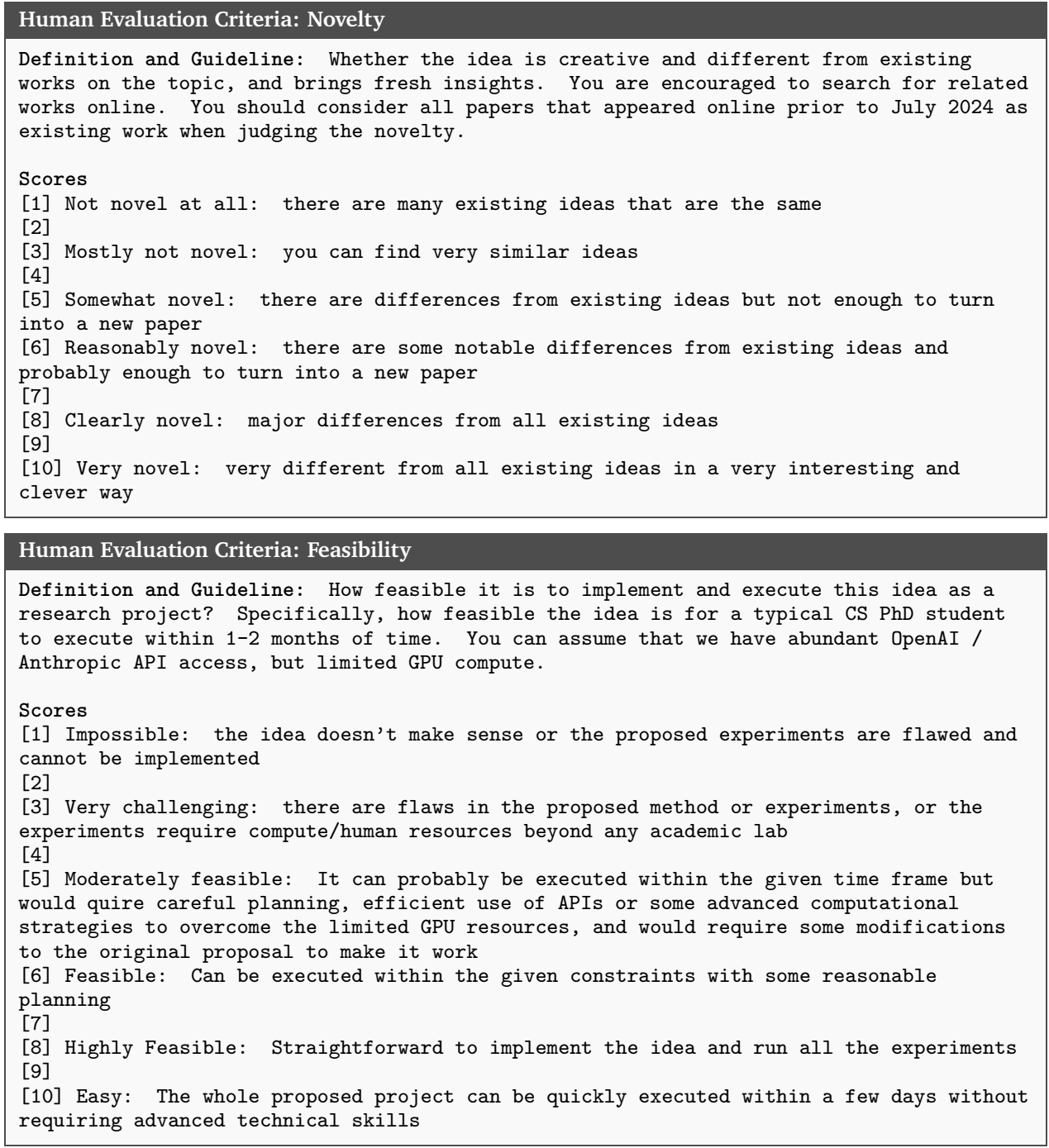

\begin{promptbox}{Human Evaluation Criteria: Novelty}
\textbf{Definition and Guideline:}
Whether the idea is creative and different from existing works on the topic, and brings fresh insights. You are encouraged to search for related works online. You should consider all papers that appeared online prior to July 2024 as existing work when judging the novelty.
\\\\
\textbf{Scores}

[1] Not novel at all: there are many existing ideas that are the same

[2]

[3] Mostly not novel: you can find very similar ideas

[4]

[5] Somewhat novel: there are differences from existing ideas but not enough to turn into a new paper

[6] Reasonably novel: there are some notable differences from existing ideas and probably enough to turn into a new paper

[7]

[8] Clearly novel: major differences from all existing ideas

[9]

[10] Very novel: very different from all existing ideas in a very interesting and clever way
\end{promptbox}

\begin{promptbox}{Human Evaluation Criteria: Feasibility}
\textbf{Definition and Guideline:}
How feasible it is to implement and execute this idea as a research project?
Specifically, how feasible the idea is for a typical CS PhD student to execute within 1-2 months of time. You can assume that we have abundant OpenAI / Anthropic API access, but limited GPU compute.
\\\\
\textbf{Scores}

[1] Impossible: the idea doesn't make sense or the proposed experiments are flawed and cannot be implemented

[2]

[3] Very challenging: there are flaws in the proposed method or experiments, or the experiments require compute/human resources beyond any academic lab

[4]

[5] Moderately feasible: It can probably be executed within the given time frame but would quire careful planning, efficient use of APIs or some advanced computational strategies to overcome the limited GPU resources, and would require some modifications to the original proposal to make it work

[6] Feasible: Can be executed within the given constraints with some reasonable planning

[7]

[8] Highly Feasible: Straightforward to implement the idea and run all the experiments

[9]

[10] Easy: The whole proposed project can be quickly executed within a few days without requiring advanced technical skills
\end{promptbox}
\caption{
    {Human evaluation criteria.} 
    Top: The definition and guideline of novelty score. 
    Bottom: The definition and guideline of feasibility score.
}
\label{fig:human-evaluation-criteria-1-2}
\end{figure*}

\section{Evaluation Details}

\subsection{Benchmark Tasks and Metrics}
\label{appendix:benchmark-metrics}
As mentioned in Section~\ref{sec:4.1}, we adopt two benchmarks including AI Idea Bench 2025 and IdeaBench. In this section, we provide detailed descriptions of their tasks and metrics to facilitate a better understanding of the experimental results.

\paragraph{AI Idea Bench 2025.}This benchmark comprises three main tasks: idea-to-topic matching (I2T), idea-to-idea matching (I2I), and idea multiple-choice evaluation (IMCQ). The first two tasks mainly assess similarity and relevance, measuring the alignment between an idea and the topic, and between an idea and a reference idea. Both tasks leverage an LLM-simulated similarity function to assign scores on a 0–5 scale.
\begin{itemize}
    \item \textbf{I2T}: Evaluate whether the generated idea aligns with the specified topic and assess its degree of alignment with the target paper’s topic, with a LLM-based similarity function $F_{IT}^D:(I_B, T_{\text{topic}})\rightarrow[0,5]$, where measures the similarity between an idea $i\in I_B$ and the topic $T_{\text{topic}}$ of the target paper. Finally, we leverage the equation below to compute the alignment score of the current baseline B with respect to the topic:
    \begin{equation}
        S_{IT}=\max_{i_j\in I_B}F_{IT}^D(i_j,T_{\text{topic}})
    \end{equation}
    \item \textbf{I2I}: Compare the generated idea with the motivation and experimental framework of the target paper using an LLM-based similarity function $F_{2I}^D:(I_B,(M_T,E_T))\rightarrow [0,5]$, where measures the similarity between an idea $i\in I_B$ and the motivation $M_T$ and experimental framework $E_T$. Finally, we compute the score of the current baseline B using the equation below:
    \begin{equation}
        S_{I^2}=\max_{i_j\in I_B}F_{2I}^D(i_j,(M_T,E_T))
    \end{equation}
\end{itemize}
The final task IMCQ focuses on evaluating the quality of generated ideas by constructing a multiple-choice question for each query. 
Each question comprises four options, where $L_1$ and $L_2$ are drawn from influential prior work exhibiting the closest conceptual alignment with the target paper T and the $L_3$ is the paper that maintains the highest degree
of similarity to all target papers in the dataset. Finally, the baseline-generated idea is incorporated as the answer option $L_4$ to form the complete option set $C=\{L_1,L_2,L_3,L_4\}$. 
The answers of each question $R_B = \{r_1, r_2, \ldots, r_n\}$ produced by baseline $B$ are compared with the correct option set $A_c$ (corresponding to baseline's idea) to compute an accuracy-based score:
\begin{equation}
    S_M=\begin{cases}
        1, &\text{if } A_c\cap R_B\not= \varnothing,\\
        0,&\text{otherwise.}
    \end{cases}
\end{equation}
The selected option is regarded as the best among the four candidates. A higher accuracy indicates that our idea is selected more frequently, reflecting stronger capability of the baseline.

\begin{figure*}[!h]
\begin{promptbox}{Human Evaluation Criteria: Excitement}
\textbf{Definition and Guideline:}
How likely the proposed idea is going to work well (e.g., better than existing baselines).
\\\\
\textbf{Scores}

[1] Poor: You cannot identify the contributions of this idea, or it's not interesting at all and you would fight to have it rejected at any major AI conference

[2]

[3] Mediocre: this idea makes marginal contributions and is very incremental

[4]

[5] Leaning negative: it has interesting bits but overall not exciting enough

[6] Learning positive: exciting enough to be accepted at a major AI conference, but still has some weaknesses or somewhat incremental

[8] Exciting: would deepen the community's understanding or make major progress in this research direction

[9]

[10] Transformative: would change the research field profoundly and worth a best paper award at major AI conferences

\end{promptbox}

\begin{promptbox}{Human Evaluation Criteria: Expected Effectiveness}
\textbf{Definition and Guideline:}
How feasible it is to implement and execute this idea as a research project?
Specifically, how feasible the idea is for a typical CS PhD student to execute within 1-2 months of time. You can assume that we have abundant OpenAI / Anthropic API access, but limited GPU compute.
\\\\
\textbf{Scores}

[1] Extremely Unlikely: The idea has major flaws and definitely won't work well

[2]

[3] Low Effectiveness: The idea might work in some special scenarios but you don't expect it to work in general

[4]

[5] Somewhat ineffective: There might be some chance that the proposed idea can work better than existing baselines but the improvement will be marginal or inconsistent

[6] Somewhat effective: There is a decent chance that the proposed idea can beat existing baselines by moderate margins on a few benchmarks

[7]

[8] Probably Effective: The idea should offer some significant improvement over current methods on the relevant benchmarks

[9]

[10] Definitely Effective: You are very confident that the proposed idea will outperform existing methods by significant margins on many benchmarks
\end{promptbox}
\caption{
    {Human evaluation criteria.} 
    Top: The definition and guideline of excitement score.    
    Bottom: The definition and guideline of expected effectiveness score.
}
\label{fig:human-evaluation-criteria-3-4}
\end{figure*}

\paragraph{IdeaBench.} This benchmark provides two types of metrics: similarity and Insight Score, with the latter assessing novelty and feasibility. The similarity-based metrics comprise two components: semantic similarity, evaluated using the F1 score of BERTScore with DeBERTa-xlarge-MNLI, and idea overlap, which measures the overlap between the generated idea and the abstract of the target paper on a 0–10 scale, accompanied by an explanation. Notably, the original paper reports a practical upper limit of 0.718 for semantic similarity.

The Insight Score is evaluated from a ranking perspective, which combined $m$ ideas from target papers and $n$ generated ideas from baselines. For each query, we leverage an LLM to rank the n generated ideas together with one target idea. If all generated ideas are ranked higher than the target idea, we assign $r_{\text{target}_i \mid q} = n + 1$. Conversely, if the target idea outperforms all generated ideas, its rank is $r_{\text{target}_i \mid q} = 1$. Otherwise, we use its actual ranking position to compute the Insight Score:
\begin{equation}
    I(LLM,q)=\frac{1}{m}\sum_{i=1}^m\frac{r_{\text{target}_i|q}-1}{n}
\end{equation}
Where the novelty and feasibility Insight Scores depend on the specific evaluation prompt.
During benchmark evaluation, to ensure stable scoring results and reduce randomness in model outputs, we set the LLM judge's decoding temperature to 0.2.

\subsection{Reward Metrics} 
\label{appendix:grm-metric}
In our FlowPIE, we primarily use novelty and feasibility as reward metrics in the generative reward model, which assigns 1–5 scale scores with a reasoning chain-of-thought to provide interpretable justifications. Novelty measures the degree to which a generated idea introduces new scientific concepts or combinations beyond the existing literature, reflecting its potential to contribute original innovations. Feasibility evaluates whether the proposed idea is technically implementable and logically consistent given current technological capabilities. 
These two metrics are widely adopted in prior research on automated scientific idea generation such as SCIPIP~\citep{wang2024scipip} and Research Agent~\citep{baek2025researchagent} also use novelty and feasibility as key criteria to assess the quality of ideas, balancing originality with practical viability. The detailed reward criterias and  LLM evaluation prompts are provided in Figure~\ref{prompt:grm}.

\subsection{Human Evaluation}
\label{appendix:human-evaluation}

\paragraph{Evaluation Criteria.}
We follow the human evaluation protocol for LLM-generated ideas proposed by~\citet{sican}, adopting a four-dimensional evaluation framework that includes novelty, feasibility, excitement, and expected effectiveness. We provide detailed evaluation guideline and criteria for human experts, including metric definitions and the interpretation of each score on the 1–10 scale, as shown in Figure~\ref{fig:human-evaluation-criteria-1-2} and~\ref{fig:human-evaluation-criteria-3-4}.

\paragraph{Details of Human Experts.} We recruit three PhD students majoring in Computer Science, each of whom has published papers in top-tier AI conferences and has served as a reviewer, ensuring the expertise and reliability of the evaluation. We provide each evaluator with anonymized ideas in random order to ensure that they are unaware of their sources, thereby enabling a fair comparison. 

\section{More Analysis}

\subsection{Effects of Different LLM Backbones}
\label{appendix:effects-of-backbones}
As shown in Table~\ref{tab:different-llms}, we conduct additional experiments on AI Idea Bench 2025 using different backbone models, including Qwen2.5-7B-Instruct (released in September 2024) and LLaMA3.1-8B-Instruct (released in July 2024). We intentionally avoid using the most recent open-source models (e.g., the Qwen3.5 series) because of potential data leakage in scientific idea generation benchmarks, where model pretraining data may include papers from the test set. The consistently superior results across different backbone models demonstrate that FlowPIE is compatible with diverse LLMs. Although its performance is still bounded by the capability of the underlying model, FlowPIE consistently delivers improvements over the corresponding backbone.

\begin{table}[!t]
    \scalebox{0.83}{
    \begin{tabular}{lccc}
        \toprule
        \textbf{Method}& \textbf{I2T} & \textbf{I2I} & \textbf{IMCQ} \\
        \midrule
        \multicolumn{4}{l}{\textit{Backbone: GPT-4o-mini (released July 2024)}}\\
        \midrule
        \textbf{SCIPIP} &4.18$\pm$0.662&3.68$\pm$0.383&0.464  \\
        \textbf{Research Agent} &4.08$\pm$0.675&3.62$\pm$0.555&0.504 \\
        \textbf{FlowPIE (\textit{Ours})}&\textbf{4.18$\pm$0.544}&\textbf{4.15$\pm$0.333}&\textbf{0.708}\\
        \midrule
        \multicolumn{4}{l}{\textit{Backbone: Qwen2.5-7B-Instruct (released Sep. 2024)}}\\
        \midrule
        \textbf{SCIPIP} &4.02$\pm$0.781&3.80$\pm$0.434&0.463 \\
        \textbf{Research Agent} &3.98$\pm$0.539&3.76$\pm$0.412&0.442 \\
        \textbf{FlowPIE (\textit{Ours})}&\textbf{4.05$\pm$0.530}&\textbf{3.87$\pm$0.420}&\textbf{0.508}\\
        \midrule
        \multicolumn{4}{l}{\textit{Backbone: LLaMA3.1-8B-Instruct (released July 2024)}}\\
        \midrule
        \textbf{SCIPIP} &3.37$\pm$0.613&3.82$\pm$0.393&0.452  \\
        \textbf{Research Agent} &3.26$\pm$0.725&3.85$\pm$0.314&0.433 \\
        \textbf{FlowPIE (\textit{Ours})}&\textbf{3.68$\pm$0.554}&\textbf{3.89$\pm$0.448}&\textbf{0.486}\\
        \bottomrule
    \end{tabular}}
    \caption{Performance comparison across different LLM backbones on the AI Idea Bench 2025.}
    \label{tab:different-llms}
\end{table}

\subsection{Reward-Benchmark Consistency Analysis}
\label{appendix:reward-benchmark-consistency}

To evaluate cross-benchmark consistency, we compute Kendall’s W across all metrics and assess rank correlation between average reward and AI Idea Bench 2025. We observe moderate but statistically significant inter-metric agreement (Kendall’s $W$ = 0.326, $p$-value = 0.023), indicating partial yet non-random ranking consistency across complementary evaluation dimensions. In addition, average reward exhibits strong correlation with AI Idea Bench 2025 (Kendall's tau $\tau$ = 0.8667, $p$-value = 0.0167), suggesting robust external alignment. Notably, FlowPIE consistently lies on the Pareto frontier, reflecting balanced multi-objective performance rather than metric-specific optimization.

\section{Prompts}
\label{appendix:prompts}

\paragraph{Initial Idea Population Construction.}
We provide the initial idea generation prompt, as shown in Figure~\ref{prompt:initial-idea-generation}.
\begin{figure*}[h]
\begin{promptbox}{Initial Idea Generation Prompt}
You are an expert in cross-domain innovation and idea generation. Your task is to generate a scientific idea that aligns with the user’s scientific intent. The objective is to produce ideas that are technically specific, experimentally verifiable, and potentially claim-supportive rather than abstract or purely descriptive. The proposed idea should address a clear scientific problem, produce a \textbf{measurable and verifiable technical effect}, and demonstrate \textbf{non-trivial scientific value} that could be considered meaningful in a paper peer review context.

 \textbf{User Query}:

 \{query\}

 And the following materials serve as background context and scientific inspiration.
They should not be summarized, paraphrased, or directly reused.

\textbf{Extracted technical elements}

\{technical\_elements\}

\textbf{Reference patent-based literature metadata}:

\{ref\_info\}

\textbf{Output Requirements:}
Generate one scientific idea together with a experimental validation plan.
The output must contain the following five sections:
\begin{itemize}[leftmargin=*, nosep]
    \item \textbf{(A) Core Method Description}: Briefly summarize the proposed idea and the scientific problem it addresses.
    \item \textbf{(B) Functional Principle}: Describe the key scientific and functional principle. When appropriate, include 1–2 formulas or models representing the underlying logic.
    \item \textbf{(C) Concrete Workflow}: Provide a concrete workflow describing the operation of the proposed idea.
    \item \textbf{(D) Potential Innovation Directions}: List 3–5 potential scientific direction for future work.
    \item \textbf{(E) Experimental Design}: Design a experiment plan to validate the effect of the idea compared with baselines. The design could specify: Experimental Setup~(such as Backbone Models, hyperparameter and so on), Variables~(Independent variables, Dependent variables), Evaluation Metrics and Baselines.
\end{itemize}

The idea should focus on scientific principles and effects, rather than just step-by-step engineering implementation

\textbf{Output Format}

Return only the five sections (A–E) described above. The experimental design should clearly demonstrate how the proposed idea leads to measurable scientific improvements.
\end{promptbox}
\caption{Prompt for the initial idea population generation.}
\label{prompt:initial-idea-generation}
\end{figure*}

\begin{figure*}[!h]
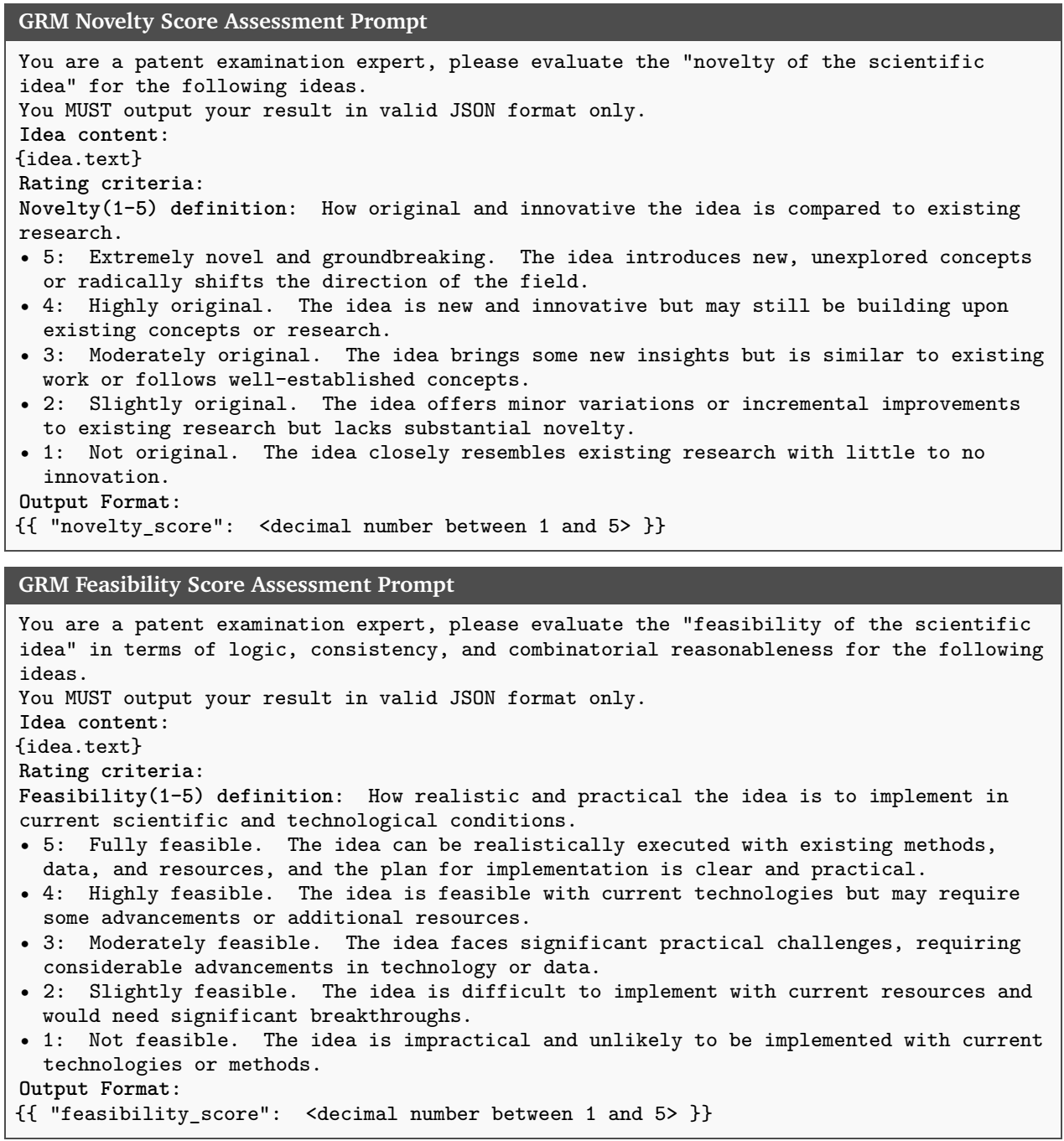

    \begin{promptbox}{GRM Novelty Score Assessment Prompt}
    You are a patent examination expert, please evaluate the "novelty of the scientific idea" for the following ideas.
    
    You MUST output your result in valid JSON format only.

    \textbf{Idea content}:
    
    \{idea.text\}

    \textbf{Rating criteria}:
    
    \textbf{Novelty(1-5) definition}: How original and innovative the idea is compared to existing research.
    \begin{itemize}[leftmargin=*, nosep]
        \item 5: Extremely novel and groundbreaking. The idea introduces new, unexplored concepts or radically shifts the direction of the field.
        \item 4: Highly original. The idea is new and innovative but may still be building upon existing concepts or research.
        \item 3: Moderately original. The idea brings some new insights but is similar to existing work or follows well-established concepts.
        \item 2: Slightly original. The idea offers minor variations or incremental improvements to existing research but lacks substantial novelty.
        \item 1: Not original. The idea closely resembles existing research with little to no innovation.
    \end{itemize}
    \textbf{Output Format}:

    \{\{
    "novelty\_score": <decimal number between 1 and 5>
    \}\}
    \end{promptbox}

    \begin{promptbox}{GRM Feasibility Score Assessment Prompt}
        You are a patent examination expert, please evaluate the "feasibility of the scientific idea" in terms of logic, consistency, and combinatorial reasonableness for the following ideas.
    
    You MUST output your result in valid JSON format only.

    \textbf{Idea content}:
    
    \{idea.text\}

    \textbf{Rating criteria}:
    
    \textbf{Feasibility(1-5) definition}: How realistic and practical the idea is to implement in current scientific and technological conditions.
    \begin{itemize}[leftmargin=*, nosep]
        \item 5: Fully feasible. The idea can be realistically executed with existing methods, data, and resources, and the plan for implementation is clear and practical.
        \item 4: Highly feasible. The idea is feasible with current technologies but may require some advancements or additional resources.
        \item 3: Moderately feasible. The idea faces significant practical challenges, requiring considerable advancements in technology or data.
        \item 2: Slightly feasible. The idea is difficult to implement with current resources and would need significant breakthroughs.
        \item 1: Not feasible. The idea is impractical and unlikely to be implemented with current technologies or methods. 
    \end{itemize}
    \textbf{Output Format}:

    \{\{
    "feasibility\_score": <decimal number between 1 and 5>
    \}\}
    \end{promptbox}
    \caption{Prompt for the generative reward model for novelty and feasibility evaluation.}
        \label{prompt:grm}
\end{figure*}

\paragraph{Parent Ideas Crossover.}
We provide the prompt for applying the crossover operator to parent ideas, as shown in Figure~\ref{prompt:crossover}.
\begin{figure*}[!h]
    \begin{promptbox}{Ideas Crossover Prompt}
    You are an expert innovation designer. Your task is to intelligently fuse two scientific ideas into one superior idea, leveraging the provided literature context including both relevant literature and diverse island literature.

    \textbf{Research Topic}:

    \{topic\}

    \textbf{Isolation Island Literature Context}:

    \{literature\_context\}

    \textbf{Scientific Idea A}(score:\{idea\_a.score\}):

    \{idea\_a.text\}

    \textbf{Scientific Idea B}(score:\{idea\_b.score\}):

    \{idea\_b.text\}

    \textbf{Requirement}:
    \begin{enumerate}[leftmargin=*, nosep]
        \item Carefully analyze both ideas and identify their strengths
        \item Study the literature context to understand existing science and identify gaps
        \item Intelligently merge the best elements from both ideas
        \item Leverage insights from island literature to enhance novelty and diversity
        \item Create a NEW, COHERENT idea that combines their advantages
        \item Ensure all five parts (A-E) are present and well-integrated
        \item The new idea should be more innovative, feasible, and diverse than either parent
    \end{enumerate}
    \textbf{Output Format}:
    \begin{itemize}[leftmargin=*, nosep]
        \item \textbf{(A) Core Method Description}(Your fused core method)
        \item \textbf{(B) Functional Principle}(Your fused functional principles)
        \item \textbf{(C) Concrete Workflow}(Your fused workflow)
        \item \textbf{(D) Potential Innovation Directions}(Your fused innovation points)
        \item \textbf{(E) Experimental Design}(Your fused experimental design)
        
    \end{itemize}
    
    \end{promptbox}
    \caption{Prompt for the parent ideas crossover operator.}
        \label{prompt:crossover}
\end{figure*}
\paragraph{Idea Mutation.}
We provide the prompt for applying the mutation operator to generated ideas, as shown in Figure~\ref{prompt:mutation}.
\begin{figure*}[h]
    \begin{promptbox}{Idea Mutation Prompt}
        You are an expert innovation improver. Your task is to enhance and mutate an existing scientific idea to make it more novel and technically sound, leveraging diverse literature information.
    
    \textbf{Research Topic}:

    \{topic\}

    \textbf{Isolation Island Literature Context}:

    \{literature\_context\}

    \textbf{Original Idea}(score:idea.score):
    
    \{idea.text\}

    \textbf{Requirement}:
    \begin{enumerate}[leftmargin=*, nosep]
        \item Identify potential weaknesses or areas for improvement
        \item Study the literature context to find inspiration from both relevant and island literature
        \item Introduce novel scientific elements or approaches inspired by diverse literature sources
        \item  Enhance the innovation level while maintaining feasibility
        \item Improve clarity and scientific depth
        \item Make sure all five parts (A-E) are strengthened
        \item Incorporate insights from island literature to increase diversity and novelty
    \end{enumerate}
    \textbf{Output Format}:
    \begin{itemize}[leftmargin=*, nosep]
        \item \textbf{(A) Core Method Description}(Your improved core method)
        \item \textbf{(B) Functional Principle}(Your improved functional principles)
        \item \textbf{(C) Concrete Workflow}(Your improved workflow)
        \item \textbf{(D) Potential Innovation Directions}(Your improved innovation points)
        \item \textbf{(E) Experimental Design}(Your improved experimental design)
        
    \end{itemize}
    \end{promptbox}
    \caption{Prompt for the parent ideas mutation operator.}
        \label{prompt:mutation}
\end{figure*}
\paragraph{Initial Idea Assessment and Fitness Evaluation using GRM.}
We provide the prompt used for novelty and feasibility assessment by the generative reward model, as shown in Figure~\ref{prompt:grm}. The fitness evaluation adopts the same prompt as that used in the GRM-based initial idea assessment, as shown in Figure~\ref{prompt:grm}.

\section{Examples}
\label{appendix:case}
We provide the full version of our demonstration of the idea in Table~\ref{case:full-demo}, DMGV, generated by FlowPIE, whose topic is improving reasoning capabilities in large language models. 
The central idea is to abstract recurrent sub-reasoning patterns into reusable reasoning macros and improve the reliability of their reuse through lightweight verification.
This directly targets a central challenge in multi-step LLM reasoning: \textit{error propagation and accumulation in intermediate reasoning steps}, while repeatedly generating similar reasoning fragments is computationally inefficient. By combining macro reuse with verifier-guided checking, the generated proposal offers a plausible direction for improving both reasoning accuracy and efficiency.

Additionally, we provide several concise summaries of generated ideas across multiple domains, particularly in Health \& Medicine, Genetics \& Molecular Biology, and Environmental Sciences, as shown in Table~\ref{example:case-study-HM},~\ref{example:case-study-GMB} and~\ref{example:case-study-ES}.
\clearpage

{\small
 
\sethlcolor{orange!30}
\renewcommand{\arraystretch}{0.96}
\setlist[itemize]{leftmargin=2em,itemsep=0pt,topsep=0pt,parsep=0pt,partopsep=0pt}

\begin{xltabular}{\textwidth}{@{}p{\dimexpr\textwidth-2\tabcolsep\relax}@{}}
\toprule
\textbf{Input topic}: \textit{The topic of this paper is improving reasoning capabilities in LLMs.} \\
\midrule
\textbf{Final Idea}: \textit{Dynamic Macro-Guided Verification (DMGV): A Modular Reasoning Augmentation Framework for LLMs} \\
\hdashline
\textit{(A) Core Concept Description}\\
Introduce a modular reasoning augmentation for LLMs called Dynamic Macro-Guided Verification (DMGV). DMGV automatically {discovers, compresses, and reuses recurrent sub-reasoning patterns (\"reasoning macros\") from LLM chain-of-thought traces}, and pairs those macros with a lightweight verifier that performs targeted, symbolic or model-based checks on macro outputs. At inference time the LLM’s generator composes higher-level proofs by invoking validated macros (rather than re-generating low-level steps), and the verifier selectively validates macro applications and issues localized repairs. The invention addresses the technical problem of error accumulation and inefficiency in multi-step LLM reasoning: by (1) turning repeated sub-proofs into compact, validated operators, and (2) constraining verification to macro interfaces, DMGV reduces cumulative hallucination, lowers tokens/computation for repeated reasoning, and yields measurable improvements in final-answer accuracy and calibration. \\
\hdashline
\textit{(B) Conceptual Functional Principle}\\
\textbf{Key principles:}
\begin{itemize}
  \item \textbf{Pattern discovery:} across many proof traces, recurrent sub-chains $f_i(x) \approx y$ (where $x$ are sub-problem inputs and $y$ their subproof outcomes) are clustered and compressed into parameterized macros $m_j(\cdot; \theta_j)$ that approximate these sub-chains with bounded approximation error $\varepsilon_j$.
  \item \textbf{Interface verification:} each macro exposes a small, checkable interface $I_j$ (inputs, claimed invariants, outputs). A verifier $V$ executes a fast check $C_j: I_j \to \{\text{pass}, \text{fail}, \text{repair-hint}\}$ that tests logical consistency or symbolic constraints cheaply.
  \item \textbf{Composition with guarded application:} the generator $G$ composes macros and raw steps. DMGV enforces that application of macro $m_j$ is only accepted if either $C_j$ passes or a bounded re-synthesis fallback is invoked.
\end{itemize}\\
\textbf{A compact formalization:}\\
Let $G$ produce a chain-of-thought trace $T = [s_1, \ldots, s_n]$. Identify subchains $S_k = s_a \ldots s_b$ that map to function $f_k : X_k \to Y_k$. Learn macro $m_j$ parameterizing a mapping $m_j(x; \theta_j) \approx f_k(x)$, with empirical error $\varepsilon_j = \mathbb{E}[\mathrm{dist}(m_j(x), f_k(x))]$. Verifier applies $C_j(x, m_j(x))$ and returns pass if $P(C_j \mid \text{correct}) \ge \tau$. Expected end-to-end error reduction when using macros: if macros are applied with reuse frequency $r$ and verifier true-positive pass rate $p$, approximate error reduction $\Delta \approx r \cdot (\varepsilon_{\text{raw}} - \varepsilon_j) \cdot p$, where $\varepsilon_{\text{raw}}$ is average raw sub-chain error without macro reuse.\\
\hdashline
\textit{(C) High-Level Conceptual Workflow}
\begin{itemize}
    \item [1.] Trace collection phase (offline or continual): Collect many generator traces $T_i$ for training problems or live tasks; annotate subchain boundaries by structural cues or automatic segmentation (e.g., argument/operation boundaries).
    \item [2.] Subchain clustering and macro synthesis: Cluster similar subchains by semantic signature (input/output shape, predicates) and distill each cluster into a macro $m_j$ with a compact parameterization and a succinct interface $I_j$.
    \item [3.] Verifier construction: For each macro, design a fast verifier $C_j$ that checks one or more invariants (e.g., arithmetic equality, type constraints, unit consistency, symbolic simplification, or shallow model prediction). Associate a verifier confidence function $\mathrm{conf}_j$.
    \item [4.] Deployment (inference): For a new query, $G$ composes a reasoning plan. When a subgoal matches a macro signature, invoke $m_j$ instead of expanding the full sub-chain; run $C_j$ on the macro’s output. If $C_j$ passes with $\mathrm{conf}_j \ge \tau$, accept and continue; if it fails or is low-confidence, fallback to re-synthesizing the sub-chain with $G$ (optionally producing a repaired macro example).
    \item [5.] Continuous update: Failed or borderline applications are used to refine $m_j$ (adjust $\theta_j$) and to improve $C_j$ thresholds; successful reuse increments reuse counters for supervisor ranking.
\end{itemize}\\
\hdashline
\textit{(D) Potential Innovation Directions}
\begin{itemize}
    \item [1.] Formalized Macro Contracts: Define formal pre/post-conditions (contracts) for macros and synthesize verifiers $C_j$ with SMT/symbolic engines for domains where symbolic checks are tractable (e.g., algebra, type systems).
    \item [2.] Cross-Domain Macro Transfer: Investigate meta-features for macro signature representation enabling reuse across tasks and modalities (textual proofs $\leftrightarrow$ code transformations $\leftrightarrow$ symbolic math).
    \item [3.] Resource-Aware Macro Selection: Add an energy/latency cost model that chooses between macro invocation and raw re-synthesis to optimize a cost-accuracy objective under latency budgets.
    \item [4.] On-Device Macro Lifecycles: Enable light-weight on-device macro discovery and caching with privacy-preserving aggregation across devices (federated macro statistics).
    \item [5.] Verifier Co-training: Co-train verifier $V$ with generator $G$ to improve verifier calibration and enable provable probabilistic guarantees (e.g., PAC-style bounds on macro error under verification).
\end{itemize}\\
\hdashline
\textit{(E) Theoretical Experimental Design}\\
\textbf{Objective:} Validate that DMGV (dynamic macro discovery + verifier-guided reuse) improves multi-step reasoning accuracy, reduces token/computation usage, and improves calibration relative to standard chain-of-thought (CoT) and self-consistency baselines.\\
\textbf{Experimental setup}
\begin{itemize}
  \item \textbf{Generator backbone:} a reasonably capable open LLM (e.g., LLaMA-2 13B or equivalent) used for generation and macro synthesis. Use same checkpoint for all methods to ensure fairness.
  \item \textbf{Verifier model:} a lightweight transformer (e.g., 200M--700M) fine-tuned for verification tasks per domain, plus domain-specific symbolic checkers where available.
  \item \textbf{Macro encoder/parameterizer:} a compact adapter (10M--50M params) which maps macro inputs to outputs.
  \item \textbf{Compute budget:} fix average allowed tokens-per-query and wall-clock time budget to model deployment constraints.
  \item \textbf{Datasets (unseen evaluation sets):} use held-out splits from multi-step reasoning benchmarks: GSM8K (arithmetic), ProofWriter (formal reasoning), StrategyQA (multi-hop), and a code-synthesis multi-step reasoning subset (e.g., problem-to-algorithm translation tasks). Use separate corpus for macro discovery (training/validation) and unseen tasks for testing to measure generalization.
\end{itemize}\\
\textbf{Independent variables}
\begin{itemize}
  \item \textbf{Method:}
  \begin{itemize}
    \item Baseline A: CoT prompting (chain-of-thought) with standard temperature (0.7).
    \item Baseline B: CoT + self-consistency (N samples, majority voting).
    \item Variant C: DMGV-static --- macros synthesized offline once from training traces, used at inference without further refinement; verifier present.
    \item Variant D: DMGV-dynamic --- macros synthesized and updated online with verifier-driven repair (full proposed method).
  \end{itemize}
  \item \textbf{Verifier fidelity (for DMGV experiments):}
  \begin{itemize}
    \item Low: verifier with conservative pass threshold $\tau_{\text{low}}$.
    \item High: verifier with higher threshold $\tau_{\text{high}}$.
  \end{itemize}
  \item \textbf{Macro reuse budget $r_{\max}$:} maximum macros allowed per query (tune values: 0, 1, 3, 10).
\end{itemize}\\
\textbf{Dependent variables / measurable indicators}
\begin{itemize}
  \item \textbf{Final answer accuracy (primary):} fraction of correct final answers on each dataset.
  \item \textbf{Token consumption per query:} average number of model tokens generated (generator + macro expansion).
  \item \textbf{Wall-clock inference latency per query.}
  \item \textbf{Chain length reduction:} average number of low-level reasoning steps synthesized vs baseline.
  \item \textbf{Verification pass rate:} fraction of macro applications accepted without fallback.
  \item \textbf{Calibration/Brier score:} calibration of model confidence on final answers.
  \item \textbf{Error localization:} proportion of failures localized to specific macro applications (for analysis).
  \item \textbf{Computational cost:} compute flops or approximate cost per query.
  \item \textbf{Statistical significance:} use paired bootstrap resampling to test differences in accuracy.
\end{itemize}\\
\textbf{Procedures}
\begin{itemize}
  \item \textbf{Macro discovery:} Use a held-out set of training problems to produce traces via $G$. Apply the DMGV clustering and synthesize macros $m_j$. For DMGV-dynamic, allow online updates during evaluation by feeding back failed macro instances; for DMGV-static, freeze library once.
  \item \textbf{Baseline runs:} Run Baseline A and B over evaluation sets, collect metrics (accuracy, tokens, latency).
  \item \textbf{DMGV runs:} Evaluate DMGV-static and DMGV-dynamic under multiple verifier fidelity settings and reuse budgets. Ensure identical inference temperature and beam settings for generator across runs.
  \item \textbf{Ablations:}
  \begin{itemize}
    \item No-verifier ablation: DMGV with macros but no verifier (accept all macros).
    \item No-compression ablation: use discovered subchains but do not compress; instead, cache and replay full subchains.
    \item Vary $r_{\max}$: measure trade-offs.
  \end{itemize}
  \item \textbf{Statistical analysis:} For each metric, compute mean $\pm$ 95\% CI across queries. Use paired statistical tests comparing DMGV-dynamic to best baseline (likely CoT+self-consistency) on accuracy and token savings.
\end{itemize}\\
\textbf{Expected measurable effects (testable hypotheses)}
\begin{itemize}
  \item \textbf{H1 (accuracy):} DMGV-dynamic increases final-answer accuracy over Baseline B by an absolute margin $\Delta_{\text{acc}} \ge 5$--$15\%$ on deep multi-step tasks (ProofWriter, GSM8K) due to reduced error accumulation in reused macros.
  \item \textbf{H2 (efficiency):} DMGV reduces average tokens per query by at least 20\% when macros are frequently reused ($r \ge 3$) versus Baseline A, while maintaining or improving accuracy.
  \item \textbf{H3 (calibration):} DMGV with verifier improves calibration (lower Brier score) because verification provides meaningful pass/fail signals that correlate with correctness.
  \item \textbf{H4 (targeted repair):} Ablations will show that verifier presence is critical: no-verifier DMGV shows token reduction but loses accuracy, demonstrating verifier's key role in maintaining correctness.
\end{itemize}\\
\textbf{Success criteria to support patent-claim-level benefit}
\begin{itemize}
  \item Statistically significant improvement in accuracy ($p < 0.05$ paired bootstrap) versus CoT+self-consistency on at least two of the four datasets.
  \item Measurable token/latency savings while meeting or exceeding baseline accuracy in deployed-cost budget scenarios.
  \item Demonstration that verifier-driven online macro refinement reduces subsequent per-macro error rate $\varepsilon_j$ by a measurable fraction (e.g., 30--60\%) across update cycles.
\end{itemize}\\
This experimental design ties the core invention --- validated, compressed reasoning macros with selective verification and repair --- to concrete, measurable improvements in accuracy, efficiency, and calibration, and includes ablations to isolate the contribution of discovery, compression, and verification.
\\
\bottomrule
\end{xltabular}
\captionof{table}{Full demonstration of an idea generated by FlowPIE.}
\label{case:full-demo}
\par}

\begin{table*}[!t]
\small
 
\sethlcolor{orange!30} 
\scalebox{0.97}{\begin{tabular}{p{\linewidth}}
\toprule
\textbf{Domain}: \textit{Health \& Medicine} \\
\midrule
\textbf{Final Idea Title}: \textit{{A Tumor-Core–Activated Therapeutic Strategy}}

\\\textbf{Summary}: The proposed strategy introduces a \hl{tumor-core–activated neutrophil reprogramming blockade} that selectively prevents infiltrating neutrophils from converging into the \textbf{pro-angiogenic dcTRAIL-R1\textsuperscript{+} T3 state} identified in the target paper. Rather than depleting neutrophils systemically or blocking VEGF broadly, this approach acts \hl{upstream at the level of fate determination}, using the hypoxic–glycolytic tumor niche as a spatial trigger to locally release an inhibitor against a critical epigenetic or transcriptional effector of the T3 program. The central advantage is that it \hl{suppresses tumor-promoting neutrophil plasticity with spatial precision while preserving peripheral neutrophil host-defense functions}, thereby overcoming the major limitations of pan-neutrophil depletion and conventional anti-angiogenic therapy. Conceptually, the strategy combines \textit{(1) microenvironment-selective activation}, ensuring that the payload is engaged primarily within the tumor core; \textit{(2) neutrophil fate interception}, blocking the chromatin remodeling and transcriptional induction required for dcTRAIL-R1 and VEGF$\alpha$ upregulation; and \textit{(3) functional antitumor consequence}, reducing T3 accumulation, intratumoral angiogenesis, and tumor growth. By targeting the newly defined deterministic reprogramming axis rather than only its downstream effector output, this framework offers a \hl{more mechanism-based, selective, and potentially safer therapeutic paradigm} for modulating tumor-associated neutrophils in solid cancers.
\\
\bottomrule
\end{tabular}}
\caption{Demonstration of an idea generated by FlowPIE in the Health \& Medicine domain.}
\label{example:case-study-HM}
\end{table*}

\begin{table*}[!t]
\small
 
\sethlcolor{orange!30} 
\scalebox{0.97}{\begin{tabular}{p{\linewidth}}
\toprule
\textbf{Domain}: \textit{Genetics \& Molecular Biology} \\
\midrule
\textbf{Final Idea Title}: \textit{{A system-and-method for "sticker-resolved flow-activation spectroscopy" of biomolecular condensates}}

\\\textbf{Summary}: The proposed framework introduces \hl{sticker-resolved flow-activation spectroscopy (SRFAS)}, a combined experimental–computational system that quantitatively separates the \textbf{energetic barrier governing condensate network rearrangement} from the \textbf{chain-length–dependent kinetic prefactors that control relaxation dynamics}. Instead of relying on conventional rheology that conflates binding energetics with polymer architecture, SRFAS integrates \hl{temperature-dependent microrheology, sequence-tunable transient probe tracers, and Bayesian inversion of coarse-grained transient-network models} to independently infer two key parameters: the flow activation energy $(E_a)$, reflecting sticker binding free energies, and the kinetic prefactor $(\tau_0(N,v,c))$, reflecting chain length, valence, and network topology. This separation enables \hl{direct recovery of molecular-scale interaction energetics from purely bulk and tracer-level rheological observables}, providing predictive control over macroscopic properties such as viscosity, relaxation time, and molecular diffusion in biomolecular condensates. By coupling experimentally measurable viscoelastic spectra, tracer diffusion, and residence-time distributions with model-based inference, the system delivers \hl{experimentally verifiable, sequence-to-rheology mapping} that surpasses conventional microrheological approaches in both parameter identifiability and predictive accuracy. As a result, SRFAS establishes a general platform for \hl{rational design, screening, and mechanistic analysis of biomolecular condensates and associative polymer materials}, with applications ranging from understanding disease-related condensate mutations to engineering synthetic phase-separating biomaterials with programmable dynamic properties.
\\
\bottomrule
\end{tabular}}
\caption{Demonstration of an idea generated by FlowPIE in the Genetics \& Molecular Biology domain.}
\label{example:case-study-GMB}
\end{table*}

\begin{table*}[!t]
\small
 
\sethlcolor{orange!30} 
\scalebox{0.97}{\begin{tabular}{p{\linewidth}}
\toprule
\textbf{Domain}: \textit{Environmental Sciences} \\
\midrule
\textbf{Final Idea Title}: \textit{{A transferable Diagnostic-and-Prediction system for ecosystem functional loss under rare short-term droughts}}

\\\textbf{Summary}: The proposed \hl{DP-Extreme (Diagnostic-and-Prediction system for Extreme drought impacts)} introduces a transferable framework that quantitatively predicts ecosystem productivity loss under \textbf{rare short-term drought events} and guides optimized mitigation decisions. Unlike conventional precipitation–productivity models that assume linear sensitivity to rainfall deficits, DP-Extreme explicitly incorporates \hl{drought rarity (return period), pre-drought resource stocks (soil moisture and nutrients), and community resilience traits (biodiversity and rooting depth)} into a hierarchical scaling law that captures the nonlinear sensitivity of aboveground net primary production (ANPP) to extreme pulses. By combining standardized pulse-exclusion experiments across environmental gradients with Bayesian parameter inference, the system estimates site-specific parameters that generate an \textbf{Extreme Sensitivity Index (ESI)}—a predictive indicator of expected ANPP loss for rare drought scenarios (e.g., 1-in-100-year events). This enables \hl{threshold-triggered mitigation strategies} such as targeted irrigation or grazing control that activate only when predicted loss exceeds critical levels. The key advantage is that DP-Extreme \hl{decouples drought rarity effects from ecological buffering traits}, producing substantially more accurate forecasts of ecosystem productivity loss and enabling resource-efficient interventions. In validation experiments, the system is designed to demonstrate \hl{significantly lower prediction error and improved decision efficiency} compared with conventional climate-only or linear sensitivity models, establishing a practical decision-support platform for managing ecosystem resilience under increasingly frequent extreme climate events.
\\
\bottomrule
\end{tabular}}
\caption{Demonstration of an idea generated by FlowPIE in the Environmental Sciences domain.}
\label{example:case-study-ES}
\end{table*}

\end{document}